\documentclass[journal,comsoc]{IEEEtran}
\usepackage[T1]{fontenc}

\ifCLASSINFOpdf
\else
\fi
\usepackage{amsmath}
\usepackage[cmintegrals]{newtxmath}
\newcommand{\etal}{\textit{et al}.~}
\newcommand{\ie}{\textit{i}.\textit{e}.~}
\newcommand{\ieno}{\textit{i}.\textit{e}.}
\newcommand{\eg}{\textit{e}.\textit{g}.~}
\newcommand{\egno}{\textit{e}.\textit{g}.} 
\newcommand{\etc}{\textit{etc}.}
\newcommand{\etcno}{\textit{etc}} 
\newcommand{\ourloss}{restitution loss }
\newcommand{\ourlossno}{restitution loss}
\newcommand{\city}    {Cityscapes}
\newcommand{\cityfog} {Foggy Cityscapes} 
\newcommand{\kit} {KITTI} 
\usepackage{pifont}
\newcommand{\std}[1]{\scriptsize$\pm${#1}}

\usepackage{makecell}
\usepackage{rotating}
\usepackage{algorithm}
\usepackage{algorithmicx}
\usepackage{algpseudocode}
\usepackage{booktabs}       
\usepackage{multirow}
\usepackage{amsmath}
\usepackage{accents}
\usepackage{cite}
\usepackage[colorlinks,linkcolor=red]{hyperref}

\makeatletter
\def\wideubar{\underaccent{{\cc@style\underline{\mskip10mu}}}}
\def\Wideubar{\underaccent{{\cc@style\underline{\mskip8mu}}}}
\makeatother
\usepackage{colortbl}
\usepackage{graphicx}
\usepackage{float}
\usepackage{subfig}
\usepackage{graphicx}
\newcommand{\tablestyle}[2]{\setlength{\tabcolsep}{#1}\renewcommand{\arraystretch}{#2}\centering\footnotesize}

\makeatletter
\def\widebar{\accentset{{\cc@style\underline{\mskip10mu}}}}
\def\Widebar{\accentset{{\cc@style\underline{\mskip8mu}}}}
\makeatother

\usepackage{enumitem}
\usepackage{caption}
\usepackage{comment}
\usepackage{xcolor}
\usepackage{ragged2e} 

\begin{document}
%
\title{Style Normalization and Restitution for Domain Generalization and Adaptation}
%
%
%
%

\author{Xin Jin, ~Cuiling Lan,~ \IEEEmembership{Member,~IEEE}, ~Wenjun Zeng,~ \IEEEmembership{Fellow,~IEEE},       ~Zhibo~Chen,~\IEEEmembership{Senior~Member,~IEEE}
	\thanks{Xin Jin and Zhibo Chen are with University of Science and Technology of China, Hefei, Anhui, 230026, China, (e-mail: chenzhibo@ustc.edu.cn)}
	\thanks{Cuiling Lan and Wenjun Zeng are with Microsoft Research Asia, Building 2, No. 5 Dan Ling Street, Haidian District, Beijing, 100080, China, (e-mail: \{culan, wezeng\}@microsoft.com)}
	\thanks{Corresponding authors: Cuiling Lan and Zhibo Chen}
	\thanks{This work was done when Jin Xin was an intern at Microsoft Research Asia.}
}

\markboth{IEEE Transactions on Multimedia}%
{Shell \MakeLowercase{\textit{et al.}}: Bare Demo of IEEEtran.cls for IEEE Communications Society Journals}
%



\maketitle

\begin{abstract}
For many practical computer vision applications, the learned models usually have high performance on the datasets used for training but suffer from significant performance degradation when deployed in new environments, where there are usually style differences between the training images and the testing images. For high-level vision tasks, an effective domain generalizable model is expected to be able to learn feature representations that are both generalizable and discriminative. In this paper, we design a novel Style Normalization and Restitution module (SNR) to simultaneously ensure both high generalization and discrimination capability of the networks. In the SNR module, particularly, we filter out the style variations (\eg, illumination, color contrast) by performing Instance Normalization (IN) to obtain style normalized features, where the discrepancy among different samples and domains is reduced. However, such a process is task-ignorant and inevitably removes some task-relevant discriminative information, which could hurt the performance. To remedy this, we propose to distill task-relevant discriminative features from the residual (\ieno, the difference between the original feature and the style normalized feature) and add them back to the network to ensure high discrimination. Moreover, for better disentanglement, we enforce a dual \ourloss constraint in the restitution step to encourage the better separation of task-relevant and task-irrelevant features. We validate the effectiveness of our SNR on different computer vision tasks, including classification, semantic segmentation, and object detection. Experiments demonstrate that our SNR module is capable of improving the performance of networks for domain generalization (DG) and unsupervised domain adaptation (UDA) on many tasks.
\end{abstract}


%
\IEEEpeerreviewmaketitle

\maketitle

\IEEEdisplaynontitleabstractindextext

%
\IEEEpeerreviewmaketitle

\section{Introduction}\label{sec:introduction}
\begin{figure}
  \centerline{\includegraphics[width=1.0\linewidth]{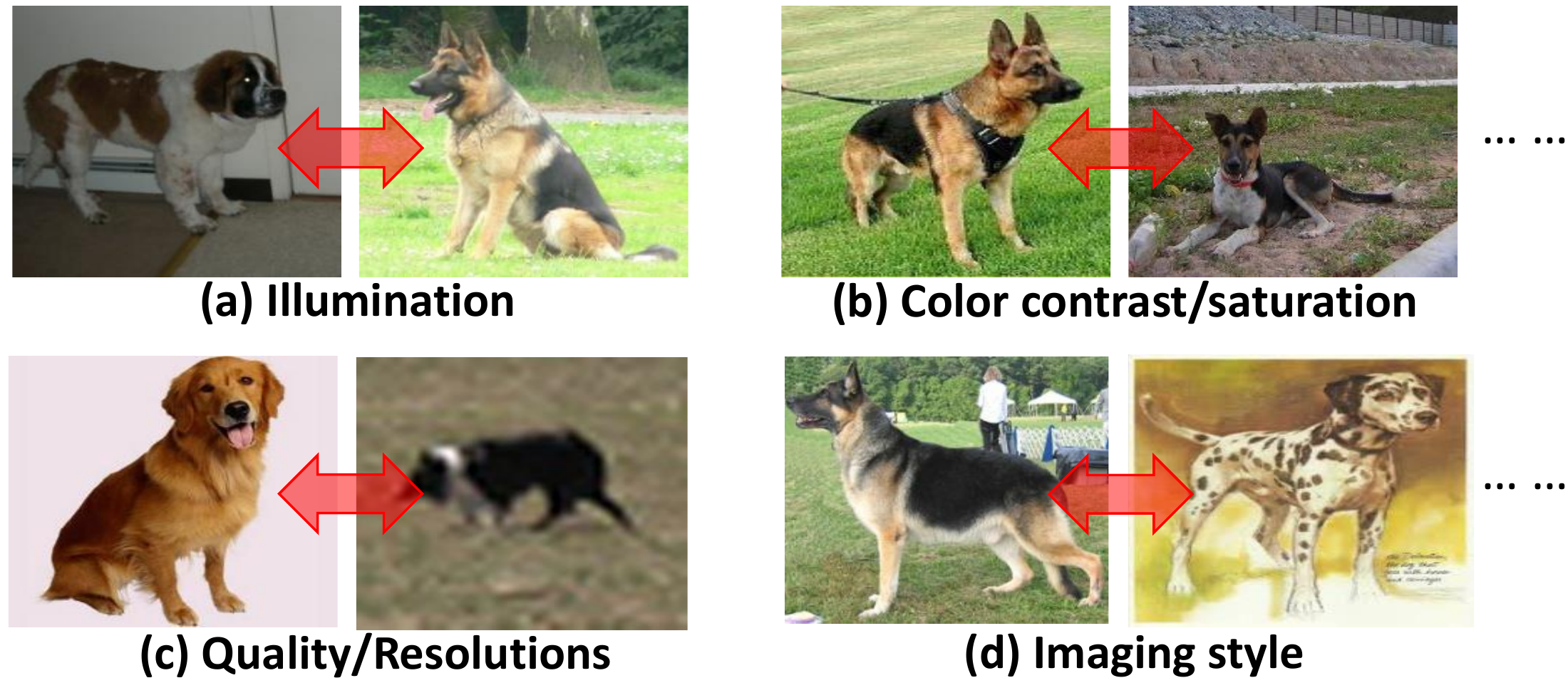}}
  \caption{Due to the differences in environments (such as lighting/camera/place/weather), the captured images present style discrepancy, such as the illumination, color contrast/saturation, quality, imaging style. These result in domain gaps between the training and testing data.}
\label{fig:examples}
\vspace{-5mm}
\end{figure}

%
%
%
%
\IEEEPARstart{D}{eep} neural networks (DNNs) have advanced the state-of-the-arts for a wide variety of computer vision tasks. The trained models typically perform well on the test/validation dataset which follows similar characteristics/distribution as the training data, but suffer from significant performance degradation (poor generalization capability) on unseen datasets that may present different styles \cite{krizhevsky2012imagenet,long2016unsupervised,ma2019deep}. This is ubiquitous in practical applications. For example, we may want to deploy a trained classification or detection model in unseen environments, like a newly opened retail store, or a house. The captured images in the new environments in general present style discrepancy with respect to the training data, such as illumination, color contrast/saturation, quality, \etc~ (as shown in Fig.~\ref{fig:examples}). These result in domain gap/shift between the training and testing. 


To address such domain gap/shift problems, many investigations have been conducted and they could be divided into two categories: domain generalization (DG) \cite{muandet2013domain,li2017deeper,shankar2018generalizing,li2018learning,carlucci2019domain,li2019episodic} and unsupervised domain adaptation (UDA) \cite{pan2009survey,ganin2014unsupervised,long2015learning,long2016unsupervised,saito2018maximum,hoffman2018cycada,peng2019moment,xu2019larger,wang2019transferable}. DG and UDA both aim to bridge the gaps between source and target domains. DG exploits only labeled source domain data while UDA can also access/exploit the unlabeled data of the target domain for training/fine-tuning. Both do not require the costly labeling on the data of target domain, which is desirable in practical applications.

\begin{figure*}
  \centerline{\includegraphics[width=1.0\linewidth]{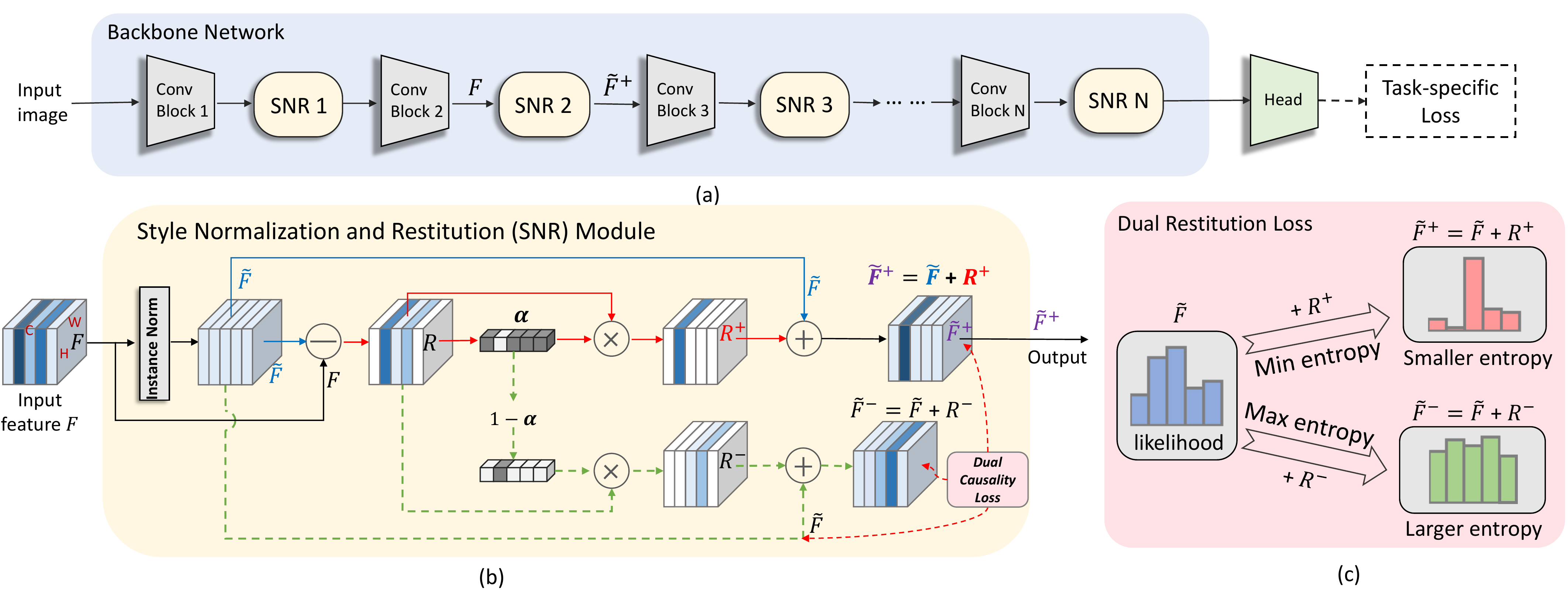}}
  \caption{Overall flowchart. (a) Our generalizable feature learning network with the proposed Style Normalization and Restitution (SNR) module being plugged in after some convolutional blocks. Here, we use ResNet-50 as our backbone for illustration. (b) Proposed SNR module. Instance Normalization (IN) is used to eliminate some style discrepancies followed by task-relevant feature restitution (marked by red solid arrows). Note the branch with dashed green line is only used for enforcing loss constraint and is discarded in inference. (c) Dual \ourloss constraint encourages the disentanglement of a residual feature $R$ to task-relevant one ($R^+$) and task-irrelevant one ($R^-$), which decreases and enhances, respectively, the entropy by adding them to the style normalized feature $\widetilde{F}$ (see Section \ref{subsec:SNR}).}
\label{fig:flowchart}
\vspace{-4mm}
\end{figure*}

In particular, due to the domain gaps, directly applying a model trained on a source dataset to an unseen target dataset typically suffers from a large performance degradation \cite{muandet2013domain,li2017deeper,shankar2018generalizing,li2018learning,carlucci2019domain,li2019episodic}. As a consequence, feature regularization based UDA methods have been widely investigated to mitigate the domain gap by aligning the domains for better transferring source knowledge to the target domain. Several methods align the statistics, such as the second order correlation \cite{sun2016return,sun2016deep,peng2018synthetic}, or both mean and variance (moment matching) \cite{zellinger2017central,peng2019moment}, in the networks to reduce the domain discrepancy on features \cite{tzeng2014deep,long2017deep}. Some other methods introduce adversarial learning which learns domain-invariant features to deceive domain classifiers \cite{ganin2014unsupervised,ganin2016domain,tzeng2017adversarial}. The alignment of domains reduces domain-specific variations but inevitably leads to loss of some discriminative information \cite{liu2019transferabletat}. Even though many works investigate UDA, the study on domain generalization (DG) is not as extensive. 

Domain generalization (DG) aims to design models that are generalizable to previously unseen domains \cite{muandet2013domain,ghifary2016scatter,motiian2017unified,shankar2018generalizing,jia2019frustratingly,song2019generalizable}, without accessing the target domain data. Classic DG approaches tend to learn domain-invariant features by minimizing the dissimilarity in features across domains \cite{muandet2013domain,motiian2017unified}. Some other DG methods explore optimization strategies to help improve generalization, \egno, through meta-learning \cite{li2018learning}, episodic training \cite{li2019episodic}, and adaptive ensemble learning \cite{zhou2020domain}. Recently, Jia~\etal \cite{jia2019frustratingly} and Zhou \etal \cite{zhou2019omni} integrate a simple but effective style regularization operation, \ieno, Instance Normalization (IN), in the networks to alleviate the domain discrepancy by reducing appearance style variations, which achieves clear improvement. However, the feature style regularization using IN is task-ignorant and will inevitably remove some task-relevant discriminative information~\cite{huang2017arbitrary,pan2018two}, and thus hindering the achievement of high performance.

In this paper, we propose a Style Normalization and Restitution (SNR) method to enhance both the generalization and discrimination capabilities of the networks for computer vision tasks. Fig.~ \ref{fig:flowchart} shows our proposed SNR module and illustrates the dual \ourloss. We propose to first perform style normalization by introducing Instance Normalization (IN) to our neural network architecture to eliminate style variations. For a feature map of an image, IN normalizes the features across spatial positions on each channel, which reserves the spatial structure but reduces instance-specific style like contrast, illumination~\cite{ulyanov2017improved,dumoulin2016learned,huang2017arbitrary}. IN reduces style discrepancy among instances and domains, but it inevitably results in the loss of some discriminative information. To remedy this, we propose to distill the task-specific information from the residues (\ieno, the difference between the original features and the instance-normalized features) and add it back to the network. Moreover, to better disentangle the task-relevant features from the residual, a dual \ourloss constraint is designed by ensuring the features after restitution of the task-relevant features to be more discriminative than that before restitution, and the features after restitution of task-irrelevant features to be less discriminative than that before restitution.

We summarize our main contributions as follows:
\begin{itemize}[leftmargin=*,noitemsep,nolistsep]

\item We propose a Style Normalization and Restitution (SNR) module, a simple yet effective plug-and-play tool, for existing neural networks to enhance their generalization capabilities. To compensate for the loss of discriminative information caused by style normalization, we propose to distill the task-relevant discriminative information from the residual (\ieno, the difference between the original feature and the instance-normalized feature).  

\item We introduce a dual \ourloss constraint in SNR to encourage the better disentanglement of task-relevant features from the residual information.


\item The proposed SNR module is generic and can be applied to various networks for different vision tasks to enhance the generalization capability, including object classification, detection, semantic segmentation, \etc. Moreover, thanks to the enhancement of generalization and discrimination capability of the networks, SNR could also improve the performance of the existing UDA networks.
\end{itemize}

Extensive experiments demonstrate that our SNR significantly improves the generalization capability of the networks and {brings improvement to the existing} unsupervised domain adaptation {networks}. This work is an extension of our conference paper \cite{jin2020style} which is specifically designed for person re-identification. In this work, we make the design generic and incorporate it into popular generic tasks, such as object classification, detection, semantic segmentation, \etcno. In addition, we tailor the dual \ourloss to these tasks by leveraging entropy comparisons. 


\section{Related Work}

\subsection{Domain Generalization (DG)} \label{subsec:relatedDG}


DG considers a challenging setting where the target data is unavailable during training. Some recent DG methods explore optimization strategies to improve generalization, \egno, through meta-learning \cite{li2018learning}, episodic training \cite{li2019episodic}, or adaptive ensemble learning \cite{zhou2020domain}. Li \etal~\cite{li2018learning} propose a meta-learning solution, which uses a model agnostic training procedure to simulate train/test domain shift during training and jointly optimize the simulated training and testing domains within each mini-batch. Episodic training is proposed in ~\cite{li2019episodic}, which decomposes a deep network into feature extractor and classifier components, and then train each component by simulating it interacting with a partner who is badly tuned for the current domain. This makes both components more robust. Zhou \etal~\cite{zhou2020domain} propose domain adaptive ensemble learning (DAEL) which learns multiple experts (for different domains) collaboratively so that when forming an \emph{ensemble}, they can leverage complementary information from each other to be more effective for an unseen target domain. Some other methods augment the samples to enhance the generalization capability~\cite{shankar2018generalizing,volpi2018generalizing}. 

Some DG approaches tend to learn domain-invariant features by aligning the domains/minimizing the feature dissimilarity across domains \cite{muandet2013domain,motiian2017unified}. Recently, several
works attempt to add Instance normalisation (IN) to CNNs to improve the model generalisation ability~\cite{pan2018two,jia2019frustratingly}. Instance normalisation (IN) layers \cite{ulyanov2016instance} could eliminate instance-specific style discrepancy~\cite{bilen2017universal} and IN has been extensively investigated in the field of image style transfer \cite{ulyanov2017improved,dumoulin2016learned,huang2017arbitrary}, where the mean and variance of IN reflect the style of images. For DG, IN alleviates the style discrepancy among domains/instances, and thus improves the domain generalization~\cite{bilen2017universal,pan2018two}. In \cite{pan2018two}, a CNN called IBN-Net is designed by inserting IN into the shallow layers for enhancing the generalization capability. However, instance normalization is task-ignorant and inevitably introduces the loss of discriminative information \cite{huang2017arbitrary,pan2018two}, leading to inferior performance. Pan \etal \cite{pan2018two} use IN and Batch Normalization (BN) together (half of channels use IN while the other half of channels use BN) in the same layer to preserve some discrimination. Nam \etal \cite{nam2018batch} determine the use of BN and IN (at dataset-level) for each channel based on learned gate parameters. It lacks the adaptivity to instances. Besides, the selection of IN or BN for a channel is hard (0 or 1) rather than soft. In this paper, we propose a style normalization and restitution module. First, we perform IN for all channels to enhance generalization. To assure high discrimination, we go a step further to consider a restitution step, which adaptively distills task-specific features from the \emph{residual} (removed information) and restitute it to the network.  

\subsection{Unsupervised Domain Adaptation (UDA)}

Unsupervised domain adaptation (UDA) belongs to a target domain annotation-free transfer learning task, where the labeled source domain data and unlabeled target domain data are available for training. Existing UDA methods typically explore to learn domain-invariant features by reducing the distribution discrepancy between the learned features of source and target domains. 
Some methods minimize distribution divergence by optimizing the maximum mean discrepancy (MMD) \cite{long2015learning,long2016unsupervised,long2017deep,yan2019weighted}, second order correlation \cite{sun2016return,sun2016deep,peng2018synthetic}, \etcno. Some other methods learn to achieve domain confusion by leveraging the adversarial learning to reduce the difference between the training and testing domain distributions \cite{ganin2014unsupervised,tzeng2017adversarial,zhang2018collaborative,zhao2018adversarial}. Moreover, some recent works tend to separate the model into feature extractor and classifier, and develop some new metric to pull close the learned source and target feature representations. In particular, Maximum Classifier Discrepancy (MCD) \cite{saito2018maximum} maximizes the discrepancy between two classifiers while minimizing it with respect to the feature extractor. Similarly, Minimax Entropy (MME)~\cite{saito2019semi} maximizes the conditional entropy on unlabeled target data w.r.t the classifier and minimizes it w.r.t the feature encoder. M3SDA \cite{peng2019moment} minimizes the moment distance among the source and target domains and per-domain classifier is used and optimized as in MCD to enhance the alignment.

Our proposed SNR module aims at enhancing the generalization ability and preserving the discrimination capability and thus enhance the existing UDA approaches. 

\subsection{Feature Disentanglement}

Learning disentangled representations can help remove irrelevant features \cite{peng2019federated}. Liu \etal introduce a unified feature disentanglement framework to learn domain-invariant features from data across different domains \cite{liu2018unified}. Lee \etal propose to disentangle the features into a domain-invariant content space and a domain-specific attributes space, producing diverse outputs without paired training data \cite{lee2018diverse}. Inspired by these works, we propose to disentangle the task-specific features from the discarded/removed residual features, in order to distill and restore the discriminative information. To encourage a better disentanglement, we introduce a dual \ourloss constraint, which enforces a higher discrimination of the feature after the restitution than  before. The basic idea is to make the class-likelihood after the restitution to be sharper than before, which enables less ambiguity of a sample.


\section{Style Normalization and Restitution}


We propose a style normalization and restitution (SNR) module which enhances the generalization capability while preserving the discriminative power of the networks for effective DG and DA. Fig.~\ref{fig:flowchart} shows the overall flowchart of our framework. Particularly, SNR can be used as a plug-and-play module for existing (\egno, classification/detection/segmentation) networks. Taking the widely used ResNet-50 \cite{he2016deep} network as an example (see Fig.~\ref{fig:flowchart}(a)), SNR module is added after each convolutional block. 

In the SNR module (see Fig.~\ref{fig:flowchart}(b)), we denote the input feature map by $F \in \mathbb{R}^{h\times w \times c}$ and the output by $\widetilde{F}^{+} \in \mathbb{R}^{h\times w \times c}$, where $h,w,c$ denote the height, width, and number of channels, respectively. We first eliminate style discrepancy among samples/instances by performing Instance Normalization (IN). Then, we propose a dedicated restitution step to distill task-relevant (discriminative) feature from the residual (previously discarded by IN, which is the difference between the original feature $F$ and the style normalized feature $\tilde{F}$), and add it to the normalized feature $\tilde{F}$. Moreover, we introduce a dual \ourloss constraint to  facilitate the better separation of task-relevant and -irrelevant features within the SNR module (see Fig.~\ref{fig:flowchart}(c)).



SNR is generic and can be used in different networks for different tasks. We also present the usages of SNR (with small variations on the dual \ourloss forms with respect to different tasks) in detail for different tasks (\ieno, object classification, detection, and semantic segmentation). Besides, since SNR can enhance the generalization and discrimination capability of networks which is also very important for UDA, SNR is capable of benefiting the existing UDA networks.

\subsection{Style Normalization and  Restitution Module} \label{subsec:SNR}



\subsubsection{Style Normalization to Reduce Domain Discrepancy} 
Real-world images could be captured by different cameras under different scenes and environments (\egno, lighting/camera/place/weather). As shown in Figure \ref{fig:examples}, the captured images present large style discrepancies (\egno, in illumination, color contrast/saturation, quality, imaging style), especially for samples from two different datasets/domains. Domain discrepancy between the source and target domain generally hinders the generalization capability of learned models. 

A learning-theoretic analysis in \cite{muandet2013domain} shows that reducing feature dissimilarity improves the generalization ability on new domains. As discussed in Section \ref{subsec:relatedDG}, Instance Normalization (IN) actually performs some kinds of style normalization which reduces the discrepancy/dissimilarity among instances/samples \cite{huang2017arbitrary,pan2018two}, so it has the power to enhance the generalization ability of networks \cite{pan2018two,jia2019frustratingly,zhou2019omni}. 

Inspired by that, in SNR module, we first try to reduce the instance discrepancy on the input feature by performing Instance Normalization \cite{ulyanov2016instance,dumoulin2016learned,ulyanov2017improved,huang2017arbitrary} as
\begin{equation}
    \begin{aligned}
        \widetilde{F} = {\rm {IN}}(F) = \gamma  (\frac{F-\mu(F)}{\sigma(F)}) + \beta,
    \end{aligned}
\end{equation}
where $\mu(\cdot)$ and $\sigma(\cdot)$ denote the mean and standard deviation computed across spatial dimensions independently for each channel and each \emph{sample/instance}, $\gamma$, $\beta$ $\in \mathbb{R}^c$ are parameters learned from the data. IN could filter out some instance-specific style information from the content. With IN performed in the feature space, Huang \etal have argued and experimentally shown that IN has more profound impacts than a simple contrast normalization and it performs a form of \emph{style normalization} by normalizing feature statistics \cite{huang2017arbitrary}.

However, IN inevitably removes some discriminative information and results in weaker discrimination capability \cite{pan2018two}. To address this problem, we propose to distill and restitute the task-specific discriminative feature from the IN removed information, by disentangling it into task-relevant feature and task-irrelevant feature with a dual \ourloss constraint (see Fig.~\ref{fig:flowchart}(b)). We elaborate on such restitution hereafter. 

\subsubsection{Feature Restitution to Preserve Discrimination} \label{sec:restitution}

As illustrated in Fig.~\ref{fig:flowchart}(b), to ensure high discrimination of the features, we propose to restitute the task-relevant feature to the network by distilling it from the residual feature $R$,
\begin{equation}
       R = F - \widetilde{F},
    \label{eq:Residual}
\end{equation}
which denotes the difference between the original input feature $F$ and the style normalized feature $\widetilde{F}$.

We disentangle the residual feature $R$ in a \emph{content adaptive} way through channel attention. This is crucial for learning generalizable feature representations since the discriminative components of different images are typically different.
Specifically, given $R$, we disentangle it into two parts: task-relevant feature $R^+ \in \mathbb{R}^{h\times w\times c}$ and task-irrelevant feature $R^- \in \mathbb{R}^{h\times w\times c}$, through masking $R$ by a learned channel attention response vector $\textbf{$\mathbf{a}$}=[a_1, a_2, \cdots, a_c] \in \mathbb{R}^c$:
\begin{equation}
    \begin{aligned}
        R^+(:,:,k) = & a_k R(:,:,k), \\ R^-(:,:,k) = & (1 - a_k)  R(:,:,k),
    \end{aligned}
    \label{eq:seperation}
\end{equation}
where $R(:,:,k) \in \mathbb{R}^{h\times w}$ denotes the $k^{th}$ channel of feature map $R$, $k=1,2,\cdots,c$. We expect the SE-like \cite{hu2018squeeze} channel attention response vector $\textbf{$\mathbf{a}$}$ to help adaptively distill the task-relevant feature for the restitution,
\begin{equation}
    \begin{aligned}
        \textbf{$\mathbf{a}$} = g(R) = \sigma({\rm W_2}\delta({\rm W_1} pool(R))),
    \end{aligned}
    \label{eq:se}
\end{equation}
where the attention module is implemented by a spatial global average pooling layer, followed by two FC layers (that are parameterized by ${{\rm W_2}} \in \mathbb{R}^{(c/r) \times c}$ and ${{\rm W_1}} \in \mathbb{R}^{c \times  (c/r)}$), $\delta(\cdot)$ and $\sigma(\cdot)$ denote ReLU activation function and sigmoid activation function, respectively. To reduce the number of parameters, a dimension reduction ratio $r$ is set to 16. 

By adding this distilled task-relevant feature $R^+$ to the style normalized feature $\widetilde{F}$, we obtain the output feature $\widetilde{F}^+$ as
\begin{equation}
       \widetilde{F}^+ = \widetilde{F} + R^+.
    \label{eq:addrelevant}
\end{equation}
Similarly, by adding the task-irrelevant feature $R^-$ to the style normalized feature $\widetilde{F}$, we obtain the contaminated feature $\widetilde{F}^-=\widetilde{F} + R^-$, which is used in the next loss optimization.

It is worth pointing out that, instead of using two independent attention modules to obtain $R^+$, $R^-$, respectively, we use $\textbf{$\mathbf{a}$}(\cdot)$, and $1-\textbf{$\mathbf{a}$}(\cdot)$ to facilitate the disentanglement. We will discuss the effectiveness of this operation in the experiment section.

\begin{figure*}
  \centerline{\includegraphics[width=1.0\linewidth]{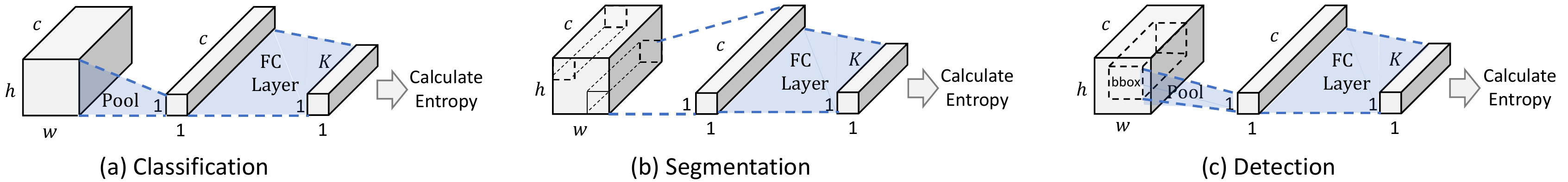}}
  \vspace{-2mm}
  \caption{Illustration of obtaining feature vector for \ourloss optimization with respect to different tasks. (a) For classification task, spatial average pooling is performed over the entire feature map ($h\times w \times c)$ to obtain a feature vector of $c$ dimensions (see Section \ref{subsubsec:cla}). (b) For segmentation task (pixel level classification), entropy is calculated for each pixel (see Section \ref{subsubsec:seg}). (c) For detection task (region level classification), spatial average pooling is performed over each groundtruth bounding box (bbox) region to obtain a feature vector of $c$ dimensions (see Section \ref{subsubsec:det}).} 
\label{fig:loss}
\vspace{-4mm}
\end{figure*}

We use the channel attention vector $\textbf{$\mathbf{a}$}$ to adaptively distill the task-relevant features for restitution for two reasons. \textbf{(a)} Those style factors (\egno, illumination, hue, contrast, saturation) are in general regarded as spatial consistent. We leverage channel attention to select the discriminative style factors distributed in different channels. \textbf{(b)} In our SNR, ``disentanglement" aims at better ``restitution" of the lost discriminative information due to Instance Normalization (IN). IN reduces style discrepancy of input features by performing normalization across spatial dimensions independently for each channel, where the normalization parameters are the same across different spatial positions. Consistent with IN, we disentangle the features and restitute the task-relevant ones to the normalized features on the channel level.

\subsubsection{Dual {Restitution} Loss Constraint}\label{sec:dual_loss}

To promote the disentanglement of task-relevant feature and task-irrelevant feature, we design a dual \ourloss constraint by comparing the discrimination capability of features \emph{before} and \emph{after} the restitution. The dual \ourloss $\mathcal{L}_{snr}$ consists of $\mathcal{L}_{SNR}^+$ and $\mathcal{L}_{SNR}^-$, \ieno, $\mathcal{L}_{SNR} = \mathcal{L}_{SNR}^+ + \mathcal{L}_{SNR}^-$. As illustrated in Figure \ref{fig:flowchart}(c), the physical meaning of the proposed dual \ourloss constraint $\mathcal{L}_{SNR}$ is that: after adding the task-relevant feature $R^+$ to the normalized feature $\widetilde{F}$, the \emph{enhanced} feature becomes more discriminative and its predicted class likelihood becomes less ambiguous (less uncertain) with a smaller entropy; on the other hand, after adding the task-irrelevant feature $R^-$ to the normalized feature $\widetilde{F}$, the \emph{contaminated} feature should become less discriminative, resulting in a larger entropy of the predicted class likelihood. 

Taking classification task as an example, we pass the spatially average pooled \emph{enhanced} feature vector $\mathbf{\widetilde{f}^+}= pool(\widetilde{F}+R^+) \in \mathbb{R}^c$ into a FC layer (of $K$ nodes, where $K$ denotes the number of classes) followed by softmax function (we denote these as $\phi(\mathbf{\widetilde{f}^+})\in \mathbb{R}^K$) and thus obtain its entropy. We denote an entropy function as $H(\cdot)=-p(\cdot) \log p(\cdot)$. Similarly, the \emph{contaminated} feature vector can be obtained by $\mathbf{\widetilde{f}^-} = pool(\widetilde{F}+R^-)$, and the style normalized feature vector is $\mathbf{\widetilde{f}} = pool(\widetilde{F})$. $\mathcal{L}_{SNR}^+$ and $\mathcal{L}_{SNR}^-$ are defined as:
\begin{align}
    \mathcal{L}_{SNR}^+ &= Softplus( H(\phi(\mathbf{\widetilde{f}^+})) - H(\phi(\mathbf{\widetilde{f}}))), 
    \label{eq:sep_loss1}
    \\ 
    \mathcal{L}_{SNR}^- &= Softplus(H(\phi(\mathbf{\widetilde{f}})) - H(\phi(\mathbf{\widetilde{f}^-}))),
    \label{eq:sep_loss2}
\end{align}
where $Softplus(\cdot) = ln(1+exp(\cdot))$ is a monotonically increasing function that aims to reduce the optimization difficulty by avoiding negative loss values. Intuitively, $\mathcal{L}_{SNR}$ {promotes better feature disentanglement of the residual $R$} for feature compensation/restitution. In {Eq.~(\ref{eq:sep_loss1})}, the loss encourages higher discrimination after restitution of the task-relevant feature by comparing the discrimination capability of features before and after the restitution. In {Eq.~(\ref{eq:sep_loss2})}, the loss encourages lower discrimination capability after restitution of the task-irrelevant feature in comparison with that before the restitution. For other tasks, \egno, segmentation, detection, there are some slight differences, \egno, in obtaining the feature vectors, which are described in the next subsection.

\subsection{Applications, Extensions, and Variants}

The proposed SNR is general. It can improve the generalization and discrimination capability of networks for DG and DA. As a plug-and-play module, SNR can be easily applied into different neural networks for different computer vision tasks, \egno, object classification, segmentation, and detection. 


As we described in Section \ref{sec:dual_loss}, we pass the spatially average pooled \emph{enhanced}/\emph{contaminated} feature vector $\mathbf{\widetilde{f}^+}/\mathbf{\widetilde{f}^-}$ into the function $H(\phi(\cdot))$ for obtaining entropy. For the different tasks of classification (\ieno, image-level classification), segmentation (\ieno, pixel level classification), detection (\ieno, region level classification), there are some differences in obtaining the feature vectors for calculating \ourlossno. Fig. \ref{fig:loss} illustrates the manners to obtain the feature vectors, respectively. We elaborate on them in the following subsections.


\subsubsection{Classification} 
\label{subsubsec:cla}

For a $K$-category classification task, we take the backbone network of ResNet-50 as an example for describing the usage of SNR. As illustrated in Fig.~\ref{fig:flowchart}(a), we could insert the proposed SNR module after each convolution block. For a SNR module, given an input feature $F$, we obtain three features--style normalized feature $\widetilde{F}$, \emph{enhanced} feature $\widetilde{F}^+$, and \emph{contaminated} feature $\widetilde{F}^-$. As shown in Fig.~\ref{fig:loss}(a), we \emph{spatially averagely pool} the features to get the corresponding feature vectors (\ieno, $\mathbf{\widetilde{f}}$, $\mathbf{\widetilde{f}^+}$, and $\mathbf{\widetilde{f}^-}$) to calculate the dual \ourloss for optimization. 

\subsubsection{Segmentation}
\label{subsubsec:seg}
Semantic segmentation predicts the label for each pixel, which is a pixel wise classification problem. Similar to classification, we insert the SNR modules to the backbone networks of segmentation. Differently, in our \ourlossno, as illustrated in Fig. \ref{fig:loss}(b), we calculate the entropy for the feature vector of each spatial position (since each spatial position has a classification likelihood) instead of over the spatially averagely pooled feature vector. To save computation and be robust to pixel noises, we take the average entropy of all pixels to calculate the \ourloss as:
\begin{equation}
    \begin{aligned}
    \mathcal{L}_{SNR}^+ &= Softplus( \frac{1}{h \times w}  \sum_{i=1}^h\sum_{j=1}^{w} H(\phi(\widetilde{F}^+(i, j, :))) \\
    &- \frac{1}{h \times w}  \sum_{i=1}^h\sum_{j=1}^{w} H(\phi(\widetilde{F}(i, j, :)))), \\ 
    \mathcal{L}_{SNR}^- &= Softplus( \frac{1}{h \times w}  \sum_{i=1}^h\sum_{j=1}^{w} H(\phi(\widetilde{F}(i, j, :))) \\
    &- \frac{1}{h \times w}  \sum_{i=1}^h\sum_{j=1}^{w} H(\phi(\widetilde{F}^-(i, j, :)))),
    \end{aligned}
    \label{eq:sep_loss}
\end{equation}
where $\widetilde{F}(i, j, :)$ denotes the feature vector of the spatial position ($i$,$j$) of the feature map $\widetilde{F}$. Note that this is slightly better than that of calculating \ourloss for each pixel in term of performance but has fewer computation.




\subsubsection{Detection} 
\label{subsubsec:det}
The widely-used object detection frameworks like R-CNN \cite{girshick2014rich}, fast/faster-RCNN \cite{ren2015faster}, mask-RCNN \cite{he2017mask}, perform object proposals, regress the bounding box of each object and predict its class, where the class prediction is based on the feature region of the bounding box. Similar to the classification task, we insert SNR modules in the backbone network. Since object detection task can be regarded as a `region-wise' (bounding box regression) classification task, as illustrated in Fig. \ref{fig:loss}(c), we calculate the entropy for each groudtruth bounding box region, with the feature vector obtained by spatially average pooling of the features within each bounding box region. We take the average entropy of all the object regions in an image to calculate the \ourlossno. 




\section{Experiment}

We validate the effectiveness and superiority of our SNR method under the domain generalization and adaptation settings for object classification (Section \ref{subsec:cla}), segmentation (Section \ref{subsec:seg}), and detection (Section \ref{subsec:det}), respectively. For each task, we describe the datasets and implementation details within each section. Moreover, without loss of generality,  we study some design choices on object classification task in Section \ref{subsec:design}. In Section \ref{subsec:visualization}, we further provide the visualization analysis.

\subsection{Object Classification}\label{subsec:cla}
We first evaluate the effectiveness of  the object classification task, under domain generalization (DG) and unsupervised domain adaptation (UDA) settings, respectively. 


\subsubsection{Datasets and Implementation Details}\label{subsec:dataset}


\begin{table*}[th]
  \centering
  \scriptsize
  \caption{Performance (in accuracy \%) comparisons with the SOTA domain generalization approaches for image classification.}
  \setlength{\tabcolsep}{2.8mm}{ 
    \begin{tabular}{l|cccc|c||cccc|c}
    \hline
    \multicolumn{1}{c|}{\multirow{2}[1]{*}{Method}} & \multicolumn{5}{c||}{PACS}            & \multicolumn{5}{c}{Office-Home} \\
\cline{2-11}          & \textit{Art}   & \textit{Cartoon}   & \textit{Photo}   & \textit{Sketch}   & \textit{Avg}   & \textit{Art}   & \textit{Clipart}   & \textit{Product}   & \textit{Real}   & \textit{Avg} \\
    \hline
    MMD-AAE  \cite{li2018domain} & 75.2  & 72.7  & \underline{96.0}  & 64.2  & 77.0  & 56.5  & 47.3  & 72.1  & 74.8  & 62.7 \\
    CCSA  \cite{motiian2017unified} & {80.5} & 76.9  & 93.6  & 66.8  & 79.4  & {59.9} & {49.9} & {74.1}  & 75.7  & {64.9} \\
    JiGen  \cite{carlucci2019domain} & 79.4  & 75.3  & {\textbf{96.2}} & 71.6  & 80.5  & 53.0  & 47.5  & 71.5  & 72.8  & 61.2 \\
    CrossGrad  \cite{shankar2018generalizing} & 79.8  & 76.8  & \underline{96.0}  & 70.2  & 80.7  & 58.4  & 49.4  & 73.9  & {75.8} & 64.4 \\
    Epi-FCR  \cite{li2019episodic} & {\underline{82.1}} & \underline{77.0} & 93.9  & {73.0} & \underline{81.5} & -     & -     & -     & -     & - \\
    
    L2A-OT~\cite{zhou2020learning} & \textbf{83.3} & \textbf{78.2} & \textbf{96.2} & \underline{73.6} & \textbf{82.8}  & \underline{60.6} & \underline{50.1} & \textbf{74.8} & \textbf{77.0} & \underline{65.6}  \\
    
    \hline
    Baseline (AGG) & 77.0  & 75.9  & \underline{96.0} & 69.2  & 79.5  & 58.9  & 49.4  & {\underline{74.3}} & {\underline{76.2}} & {64.7} \\

    SNR
   & 80.3  & {\textbf{78.2}} & 94.5  & {\textbf{74.1}} & {\underline{81.8}} & {\textbf{61.2}} & {\textbf{53.7}} & {74.2} & {75.1}  & {\textbf{66.1}} \\
   
    \hline
    \end{tabular}}%
  \label{tab:sto_dg}%
\end{table*}%

\begin{table*}[t]\centering
    \caption{Ablation study and performance comparisons (\%) with the SOTA UDA approaches for image classification.}
    \vspace{-1mm}
	\captionsetup[subffloat]{justification=centering}
	\subfloat[Results on Digit-Five. \label{tab:sto_digit5}]
	{
		\tablestyle{3.0pt}{1.0}
            \begin{tabular}{l|ccccc|c}
            \hline
            \multicolumn{1}{c|}{\multirow{2}[1]{*}{Method}} & \multicolumn{6}{c}{Digit-Five} \\
        \cline{2-7}          & \textit{mm} & \textit{mt} & \textit{up} & \textit{sv} & \textit{syn} & \textit{Avg} \\
            \hline
            DAN  \cite{long2015learning} & 63.78 & 96.31 & 94.24 & 62.45 & 85.43 & 80.44 \\
            CORAL  \cite{sun2016return} & 62.53 & 97.21 & 93.45 & 64.40 & 82.77 & 80.07 \\
            DANN  \cite{ganin2016domain} & 71.30 & 97.60 & 92.33 & 63.48 & 85.34 & 82.01 \\
            JAN  \cite{long2017deep} & 65.88 & 97.21 & 95.42 & 75.27 & 86.55 & 84.07 \\
            ADDA  \cite{tzeng2017adversarial} & 71.57 & 97.89 & 92.83 & 75.48 & 86.45 & 84.84 \\
            DCTN  \cite{xu2018deep} & 70.53 & 96.23 & 92.81 & 77.61 & 86.77 & 84.79 \\
            MEDA  \cite{wang2018visual} & 71.31 & 96.47 & 97.01 & 78.45 & 84.62 & 85.60 \\
            MCD  \cite{saito2018maximum} & 72.50 & 96.21 & 95.33 & 78.89 & 87.47 & 86.10 \\
            M3SDA  \cite{peng2019moment} & 69.76 & \underline{98.58} & 95.23 & 78.56 & 87.56 & 86.13 \\
            M3SDA-$\beta$  \cite{peng2019moment} & \underline{{72.82}} & 98.43 & \underline{96.14} & \underline{81.32} & \underline{89.58} & \underline{87.65} \\
            \hline
            
            
            

            
            
            Baseline~(M3SDA) & 69.76 & \underline{98.58} & 95.23 & 78.56 & 87.56 & 86.13 \\
            
            
                
                 
                 
            SNR-M3SDA & {\textbf{83.40}}
                & {\textbf{99.47}}
                & {\textbf{98.82}}
                & {\textbf{91.10}}
                & {\textbf{97.81}}
                & {\textbf{94.12}} \\

            \hline
            \end{tabular}%
    }
    \hspace{2mm}
	\captionsetup[subfloat]{captionskip=4pt}
	\subfloat[Results on mini-DomainNet. \label{tab:sto_mini_domainNet}]
	{
		\tablestyle{3.5pt}{1.39}
            \begin{tabular}{l|cccc|c}
            \hline
            \multicolumn{1}{c|}{\multirow{2}[1]{*}{Method}} & \multicolumn{5}{c}{mini-DomainNet} \\
        \cline{2-6}          & \textit{clp} & \textit{pnt} & \textit{rel} & \textit{skt} & \textit{Avg} \\
            \hline
            MCD  \cite{saito2018maximum} & 62.91 & 45.77 & 57.57 & 45.88 & 53.03 \\
            DCTN  \cite{xu2018deep} & 62.06 & 48.79 & 58.85 & 48.25 & 54.49 \\
            DANN   \cite{ganin2016domain} & 65.55 & 46.27 & 58.68 & 47.88 & 54.60 \\
            M3SDA  \cite{peng2019moment} & 64.18 & 49.05 & 57.70 & {49.21} & 55.03 \\
            
            M3SDA-$\beta$  \cite{peng2019moment} & 65.58 & \underline{50.85} & 58.40 & \underline{49.33} & \underline{56.04} \\            
            MME  \cite{saito2019semi} & \underline{{68.09}} & 47.14 & {\textbf{63.33}} & 43.50 & {55.52} \\
            
            
            
            \hline
            
            Baseline (M3SDA) & 64.18 & 49.05 & 57.70 & {49.21} & 55.03 \\     
            


            SNR-M3SDA & {\textbf{66.81}}
            & {\textbf{51.25}}
            & \underline{60.24}
            & {\textbf{53.98}}
            & {\textbf{58.07}} \\

            
            \hline
            \end{tabular}%
	}
	\label{tab:sto_uda}
	\vspace{-4mm}
\end{table*}

\begin{table*}[htbp]
  \centering
  \scriptsize
  \caption{Performance (\%) comparisons with the SOTA approaches for UDA on the full DomainNet dataset.}
  \setlength{\tabcolsep}{5.6mm}{ 
        \begin{tabular}{l | c c c c c c | c}
        \hline
        \multicolumn{1}{c|}{\multirow{2}[1]{*}{Method}} & \multicolumn{6}{c|}{DomainNet} \\
        \cline{2-8}
        & \textit{clp} & \textit{inf} & \textit{pnt} & \textit{qdr} & \textit{rel} & \textit{skt} & \textit{Avg} \\
        \hline
        
        MCD  \cite{saito2018maximum} & -- & -- & -- & -- & -- & -- & 38.51 \\
            
        DCTN  \cite{xu2018deep} & -- & -- & -- & -- & -- & -- & 38.27 \\
        
        DANN \cite{ganin2016domain} & 45.5\std{0.59} & 13.1\std{0.72} & 37.0\std{0.69} & 13.2\std{0.77} & 48.9\std{0.65} & 31.8\std{0.62} & 32.65 \\
        
        DCTN~\cite{xu2018deep} & 48.6\std{0.73} & 23.5\std{0.59} & 48.8\std{0.63} & 7.2\std{0.46} & 53.5\std{0.56} & 47.3\std{0.47} & 38.27 \\
        MCD~\cite{saito2018maximum} & 54.3\std{0.64} & 22.1\std{0.70} & 45.7\std{0.63} & 7.6\std{0.49} & 58.4\std{0.65} & 43.5\std{0.57} & 38.51 \\
        
        
        Baseline (M3SDA) & 58.6\std{0.53} & 26.0\std{0.89} & 52.3\std{0.55} & 6.3\std{0.58} & 62.7\std{0.51} & 49.5\std{0.76} & 42.67 \\
        
        SNR-M3SDA & \textbf{63.8}\std{0.22} & \textbf{27.6}\std{0.27} & \textbf{54.5}\std{0.14} & \textbf{15.8}\std{0.29} & \textbf{63.8}\std{0.37} & \textbf{54.5}\std{0.21} & \textbf{46.67} \\
        \hline
        \end{tabular}

}%
  \label{tab:full_domainnet}%
  \vspace{-2mm}
\end{table*}%

We conduct experiments on four classification datasets of multiple domains: PACS (includes \textit{Sketch}, \textit{Photo}, \textit{Cartoon}, and \textit{Art}), Office-Home \cite{venkateswara2017Deep}, Digit-Five (indicates five most popular digit datasets, MNIST \cite{lecun1998mnist}, MNIST-M \cite{ganin2015unsupervised}, USPS \cite{hull1994database}, SVHN \cite{netzer2011svhn}, Synthetic \cite{ganin2015unsupervised}), and DomainNet \cite{peng2019moment}. The detailed datasets introduction and experimental implementation details can be found in \textbf{Supplementary}.

\subsubsection{Results on Domain Generalization}

DG is very attractive in practical applications, which aims at ``train once and
run everywhere”. 
We perform experiments on PACS and Office-Home for
DG. There are very few works in this field. \textbf{MMD-AAE}~\cite{li2018domain} learns a domain-invariant embedding by minimizing the Maximum Mean Discrepancy (MMD) distance to align the feature representations. \textbf{CCSA}~\cite{motiian2017unified} proposes a semantic alignment loss to reduce the feature discrepancy among domains. \textbf{CrossGrad}~\cite{shankar2018generalizing} uses domain discriminator to guide the data augmentation with adversarial gradients. \textbf{JiGen}~\cite{carlucci2019domain} jointly optimizes object classification and the Jigsaw puzzle problem. \textbf{Epi-FCR}~\cite{li2019episodic} leverages episodic training strategy to simulate domain shift during the model training. L2A-OT~\cite{zhou2020learning} synthesizes source domain training data by using data augmentation techniques, which explicitly increases the diversity of available training domains and leads to a generalizable model. 

Table \ref{tab:sto_dg} shows the comparisons with the state-of-the-art methods. We can see that the proposed scheme \emph{SNR} achieves the second best and the best average accuracy on PACS and Office-Home, respectively. \emph{SNR} outperforms our baseline \emph{Baseline (AGG)} that aggregates all source domains to train a single model by \textbf{2.3\%} and \textbf{1.4\%} for PACS and Office-Home, respectively. Note that, L2A-OT~\cite{zhou2020learning} additionally employs a data generator to synthesize data to augment the source domains, and thus increasing the diversity of available training domains, leading to a more generalizable model (best performance on PACS). {Our method, which reduces the discrepancy among different samples while ensuring high discrimination, is conceptually complementary to L2A-OT, which increases the diversity of input to make the model robust to the input. We believe that adding SNR on top of L2A-OT would further improve the performance.}
\begin{table*}[t]
  \centering
  \caption{Effectiveness of our SNR, compared to other normalization-based methods for domain generalizable classification. Note that the \emph{italics} denotes the left-out target domain. We use ResNet18 as our backbone. 
  }
    \begin{tabular}{c|cccc|c||cccc|c}
    \toprule
    
    \multicolumn{1}{c|}{\multirow{2}[1]{*}{Method}}
    
 & \multicolumn{5}{c||}{PACS}            & \multicolumn{5}{c}{Office-Home} \\
\cline{2-11}          & \textit{Art}   & \textit{Cat}   & \textit{Pho}   & \textit{Skt}   & \textit{Avg}   & \textit{Art}   & \textit{Clp}   & \textit{Prd}   & \textit{Rel}   & \textit{Avg} \\
    \hline
    AGG   & 77.0  & 75.9  & \textbf{96.0}  & 69.2  & 79.5  & 58.9  & 49.4  & 74.3  & 76.2  & 64.7 \\
    AGG-All-IN & 78.8  & 74.9  & 95.8  & 70.2  & 79.9  & 59.5  & 49.3  & 75.1  & 76.8  & 65.2 \\
    AGG-IN & 78.9  & 75.3  & 95.4  & 70.8  & 80.1  & 59.9  & 49.9  & 74.1  & 76.7  & 65.2 \\
    
    AGG-IBN-a & 79.0	& 74.3	& 94.8	& 72.9	& 80.3	& 59.7	& 48.2	& 75.6	& 77.5	& 65.3
 \\
    
    AGG-IBN-b & 79.1  & 74.7  & 94.9  & 72.9  & 80.4  & 59.5  & 48.5  & {75.7}  & \textbf{77.8}  & 65.4 \\    
    
    AGG-All-BIN  & 79.1	& 74.2	& 95.2	& 72.5	& 80.3	& 58.9	& 48.7	& 76.2	& 77.6	& 65.4 \\
    
    AGG-All-BIN$^{*}$ & 79.8  & 74.5  & 95.4  & 72.6  & 80.6  & 59.8  & 48.9  & \textbf{75.8}  & 77.7  & 65.6  \\
    

    \hline
    SNR   & \textbf{80.3}  & \textbf{78.2}  & 94.5  & \textbf{74.1}  & \textbf{81.8}  & \textbf{61.2}  & \textbf{53.7}  & 74.2  & 75.1  & \textbf{66.1} \\
    \bottomrule
    \end{tabular}%
  \label{tab:ablation_study}%
  \vspace{-2mm}
\end{table*}%

\begin{table*}[t]
  \centering
  \caption{Ablation study on the dual \ourloss $\mathcal{L}_{SNR}$ for domain generalizable classification. Backbone is ResNet18.}
    \begin{tabular}{c|cccc|c||cccc|c}
    \toprule
    \multirow{2}[3]{*}{Method} & \multicolumn{5}{c||}{PACS}            & \multicolumn{5}{c}{Office-Home} \\
\cline{2-11}           & \textit{Art}   & \textit{Cat}   & \textit{Pho}   & \textit{Skt}   & \textit{Avg}   & \textit{Art}   & \textit{Clp}   & \textit{Prd}   & \textit{Rel}   & \textit{Avg} \\
    \hline
    Baseline (AGG)   & 77.0  & 75.9  & \textbf{96.0}  & 69.2  & 79.5  & 58.9  & 49.4  & \textbf{74.3}  & \textbf{76.2}  & 64.7 \\
    

    SNR w/o $\mathcal{L}_{SNR}$ & 79.0  & 77.2  & 93.8  & 73.1  & 80.8  & 61.2  & 51.3  & 73.9  & 74.9  & 65.3 \\
    SNR w/o $\mathcal{L}_{SNR}^+$ & 79.2  & 77.5  & 93.6  & \textbf{74.4}  & 81.2  & 61.0  & 51.4  & 73.7  & 74.6  & 65.2 \\
    SNR w/o $\mathcal{L}_{SNR}^-$ & 78.9  & 77.1  & 93.7  & 74.1  & 81.0  & \textbf{61.4}  & 51.9  & 74.0  & 75.0  & 65.6 \\
    
    SNR w/o Comparing & 78.7	& 77.7	& 93.9	& 74.3	& 81.2	& 61.1	& 51.9	& 74.1	& 74.6	& 65.4 \\ 
    
    \hline
    SNR   & \textbf{80.3}  & \textbf{78.2}  & 94.5  & 74.1  & \textbf{81.8}  & 61.2  & \textbf{53.7}  & 74.2  & 75.1  & \textbf{66.1} \\
    \bottomrule
    \end{tabular}%
  \label{tab:loss_snr}%
  \vspace{-4mm}
\end{table*}%

\subsubsection{Results on Unsupervised Domain Adaptation}

The introduction of SNR modules to the networks of existing UDA methods could reduce the domain gaps and preserve discrimination. It thus facilitates the domain adaptation.
Table \ref{tab:sto_uda} shows the experimental results on the two datasets Digit-Five and mini-DomainNet. Table~\ref{tab:full_domainnet} shows the results on the full DomainNet dataset. Here, we use the alignment-based UDA method M3SDA \cite{peng2019moment} as our baseline UDA network for domain adaptive classification. We refer to the scheme after using our SNR as \emph{SNR-M3SDA}. 


We have the following observations. 1) For the overall performance (as shown in the column marked by Avg), the scheme \emph{SNR-M3SDA} achieves the best performance on both datasets, outperforming the
second-best method (M3SDA-$\beta$~\cite{peng2019moment} significantly by 6.47\% on Digit-Five, and 2.03\% on mini-DomainNet in accuracy. 2) In comparison with the baseline scheme \emph{Baseline (M3SDA~\cite{peng2019moment})}, which uses the aligning technique in \cite{peng2019moment} for domain adaptation, the introduction of SNR (scheme \emph{SNR-M3SDA}) brings significant gains of 7.99\% on Digit-Five, 3.04\% on mini-DomainNet, and 4.0\% on full DomainNet in accuracy, demonstrating the effectiveness of SNR modules for UDA.

\subsubsection{Ablation Study}

We first perform comprehensive ablation studies to demonstrate the effectiveness of 1) the SNR module, 2) the proposed dual \ourloss constraint. We evaluate the models under the domain generalization (on PACS and Office-Home datasets) setting, with ResNet18 as our backbone network. Besides, we validate that SNR is beneficial to UDA and is complementary to the existing UDA techniques on the Digital-Five dataset.



\noindent\textbf{Effectiveness of SNR.} Here we compare several schemes with our proposed SNR. \emph{\textbf{AGG}}: a simple strong baseline that aggregates all source domains to train a single model. \emph{\textbf{AGG-All-IN}}: on top of \emph{AGG} scheme, we replace all the Batch Normalization(BN) \cite{ioffe2015batch} layers in AGG by Instance Normalization(IN). \emph{\textbf{AGG-IN}}: on top of \emph{AGG} scheme, an IN layer is added after each convolutional block/stage (the first four blocks) of backbone (ResNet18), respectively. \emph{\textbf{AGG-IBN-a}, \textbf{AGG-IBN-b}}: Following IBNNet~\cite{pan2018two}, we insert BN and IN in parallel at the beginning of the first two residual blocks for scheme \emph{AGG-IBN-a}, and we add IN to the last layers of the first two residual blocks to get \emph{AGG-IBN-b}. \emph{\textbf{AGG-All-BIN}}: following ~\cite{nam2018batch}, we replace all BN layers of the baseline network by Batch-Instance Normalization (BIN) to get the scheme \emph{AGG-All-BIN}, which uses dataset-level learned gates to determine whether to do instance normalization or batch normalization for each channel. \emph{AGG-All-BIN$^*$} denotes a variant of \emph{AGG-All-BIN}, where we replace the original dataset-level learned gates with content-adaptive gates (via channel attention layer~\cite{hu2018squeeze}) for the selection of normalization manner. \emph{\textbf{AGG-SNR}}: our final scheme where a SNR module is added after each block (of the first four convolutional blocks/stages) of backbone, respectively (see Fig. \ref{fig:flowchart}). We also refer to it as \emph{\textbf{SNR}} for simplicity. Table \ref{tab:ablation_study} shows the results. We have the following observations/conclusions:


\noindent\textbf{1)} Such normalization based methods, including \emph{AGG-All-IN}, \emph{AGG-IN}, \emph{AGG-IBN-a}, \emph{AGG-IBN-b}, \emph{AGG-BIN} and \emph{AGG-BIN$^{*}$} improve the performance of the baseline scheme \emph{AGG} by \textbf{0.4\%}, \textbf{0.6\%}, \textbf{0.8\%}, \textbf{0.9\%}, \textbf{0.8\%}, and \textbf{1.1\%} in average on PACS, respectively, which demonstrates the effectiveness of IN for improving the model generalization capability. 

\noindent\textbf{2)} \emph{AGG-All-BIN$^{*}$} outperforms \emph{AGG-All-IN} by 0.9\%/0.4\% on PACS/Office-Home. This because that IN introduces some loss of discriminative information and the selective use of BN and IN can preserve some discriminative information. \emph{AGG-All-BIN$^{*}$} slightly outperforms the original \emph{AGG-All-BIN}, demonstrating that the instance-adaptive determination of IN or BN is better than dataset-level determination (\ieno, same selection results of the use of IN and BN for all instances).



\noindent\textbf{3)} Thanks to the our compensation of the task-relevant information in the proposed restitution step, our final scheme \emph{SNR} achieves superior performance, which significantly outperforms all the baseline schemes. In particular, \emph{SNR} outperforms \emph{AGG} by 2.3\% and 1.4\% on PACS and
Office-Home, respectively. \emph{SNR} outperforms \emph{AGG-IN} by 1.7\% and 0.9\% on PACS and
Office-Home, respectively. Such large improvements also demonstrate that style normalization is not enough, and the proposed restitution is critical. Thanks to our restitution design, \emph{SNR} outperforms \emph{AGG-BIN$^{*}$} by 1.2\% and 0.5\% on PACS and Office-Home, respectively.    


\noindent\textbf{Effectiveness of Dual  Restitution Loss.}
Here, we perform ablation study on the proposed dual \ourloss constraint. Table~\ref{tab:loss_snr} shows the results. 

\noindent1) We observe that our final scheme SNR outperforms the scheme without
the dual \ourloss (\ieno, scheme \emph{SNR w/o $\mathcal{L}_{SNR}$}) by 1.0\% and 0.8\% on PACS and Office-Home, respectively. The dual \ourloss effectively promotes the disentanglement of task-relevant information and task-irrelevant information. Besides, both the constraint on the enhanced feature $\mathcal{L}_{SNR}^+$ and that on the contaminated feature $\mathcal{L}_{SNR}^-$ contribute to the good feature disentanglement. 

\noindent2) In  $\mathcal{L}_{SNR}$, we compare the entropy of the predicted class likelihood of features \emph{before} and \emph{after} the feature restitution process to encourage the distillation of discriminative features. To verify the effectiveness of this strategy, we compare it with the scheme without comparing \emph{SNR w/o Comparing}, which minimizes the entropy loss of the predicted class likelihood of the \emph{enhanced} feature $\mathbf{f^+}$ and maximizes the entropy loss of the predicted class likelihood of the \emph{contaminated} feature $\mathbf{f^-}$, \ieno, without comparison with the normalized feature. Table~\ref{tab:loss_snr} reveals that \emph{SNR} with the comparison outperforms \emph{SNR w/o Comparing} by \textbf{0.6\%} on PACS, and \textbf{0.7\%} on Office-Home.


\begin{table}[t]
  \centering
  \caption{Influence of SNR modules for DG and UDA respectively on top of a simple ResNet-50 baseline without incorporating other UDA methods. DG schemes \emph{Baseline(AGG)} and \emph{SNR} do not use target domain data for training. \emph{SNR-UDA} uses target domain unlabeled data for training.}
    \setlength{\tabcolsep}{1.0mm}{
    \begin{tabular}{c|cccccc}
    \toprule
    \multirow{2}[0]{*}{Method} & \multicolumn{6}{c}{Digit-Five} \\
\cline{2-7}           & \textit{mm}   & \textit{mt}   & \textit{up}   & \textit{sv}   & \textit{syn}  & \textit{Avg} \\
    \hline

    Baseline(AGG) & 63.37 & 90.50 & 88.71 & 63.54 & 82.44 & 77.71 \\
    
    SNR & 65.46	& 93.14	& 88.32	& 63.43	& 84.08	& 78.89 \\

    \hline
    
    SNR-UDA & 65.86	& 93.24	& 89.79	& 65.21	& 85.04	& 79.83 \\

    \bottomrule
    \end{tabular}}%
  \label{tab:SNR-UDA}%
  \vspace{-4mm}
\end{table}%


\noindent\textbf{SNR for DG and UDA.}
One may wonder how about the performance when exploiting  UDA directly, where other UDA-based methods (\egno, \emph{M3SDA}) are not used together. We perform this experiment by training the scheme \emph{SNR} (the baseline (VGG) powered by SNR modules) using source domain labeled data and target domain unlabeled data. We refer to this scheme as \emph{SNR-UDA}. Table \ref{tab:SNR-UDA} shows the comparisons on Digital-Five. The difference between \emph{SNR} and \emph{SNR-UDA} is that \emph{SNR-UDA} uses target domain unlabeled data for training while \emph{SNR} only uses source domain data. We can see that \emph{SNR-UDA} outperforms \emph{SNR} by 0.94\% in average accuracy. Moreover, as shown in Table~\ref{tab:SNR-UDA}(a), introducing SNR modules to the baseline UDA scheme \emph{M3SDA} brings 7.99\% gain for UDA. These demonstrate SNR is helpful for UDA, especially when it is jointly used with existing UDA method. SNR modules reduce the style discrepancy between source and target domains, which \emph{eases the alignment and adaptation}. Note that SNR modules reduce style discrepancy of instances for the source domain and target domain. However, there is a lack of explicit interaction between source and target domain after the resitition of discriminative features. Thus, the explicit alignment like in \emph{M3SDA} is still very useful for UDA.

\subsubsection{Design Choices of SNR}
\label{subsec:design}

\noindent\textbf{Which Stage to Add SNR?} We compare the cases of adding a single SNR module to a different convolutional block/stage, and to all the four stages (\ie, stage-1 to 4) of the ResNet18 (see Fig.~\ref{fig:flowchart}(a)), respectively. The module is added after the last layer of a convolutional block/stage. Table \ref{tab:stage} shows that on top of the baseline scheme \emph{Baseline (AGG)}, SNR is not sensitive to the inserted position and brings gain at each stage. Besides, when SNR is added to all the four stages, we achieve the best performance.


\begin{table}[th]
  \centering
  \caption{Ablation study on which stage to add SNR.}
    \setlength{\tabcolsep}{3.8mm}{
    \begin{tabular}{c|ccccc}
    \toprule
    \multirow{2}[0]{*}{Method} & \multicolumn{5}{c}{PACS} \\
\cline{2-6}           & \textit{Art}   & \textit{Cat}   & \textit{Pho}   & \textit{Skt}   & \textit{Avg}  \\
    \hline
    Baseline (AGG)   & 77.0  & 75.9  & 96.0  & 69.2  & 79.5 \\
    stage-1 & 77.5  & 76.2  & \textbf{96.2} & 69.9  & 80.0 \\
    stage-2 & 78.9  & 76.8  & 95.5  & 71.9  & 80.8 \\
    stage-3 & 80.1  & 76.5  & 95.3  & 72.4  & 81.1 \\
    stage-4 & 77.8  & 77.1  & 94.8  & 72.5  & 80.6 \\
    \textbf{stage-all} & \textbf{80.3} & \textbf{78.2} & 94.5  & \textbf{74.1} & \textbf{81.8} \\
    \bottomrule
    \end{tabular}}%
  \label{tab:stage}%
\end{table}%

\begin{figure*}[t]
  \centerline{\includegraphics[width=1.0\linewidth]{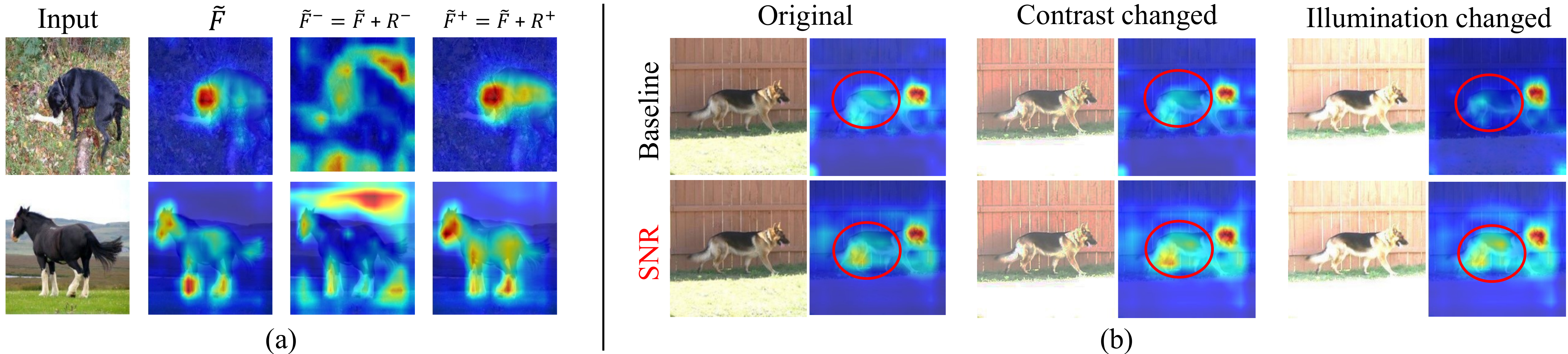}}
  \caption{(a) Activation maps of different features within an SNR module (SNR 3). They show that SNR can disentangle out the task-relevant (classification-relevant) object features well (\ieno, $R^+$). (b) Activation maps of ours (bottom) and the baseline \emph{Baseline (AGG)} (top) w.r.t images of varied styles. The maps of SNR are more consistent for images of different styles.}
\label{fig:vis_ftp}
\vspace{-4mm}
\end{figure*}

\noindent\textbf{Influence of Disentanglement Design.}
In our SNR module, as described in Eq.~(\ref{eq:seperation})(\ref{eq:se}) of Section \ref{sec:restitution}, we use the learned channel attention vector $a(\cdot)$, and its complementary one $1-a(\cdot)$ as masks to obtain task-relevant feature $R^+$ and task-irrelevant feature $R^-$, respectively. Here, we study the influence of different disentanglement designs within SNR. \textbf{\emph{SNR${_{conv}}$}}: we disentangle the residual feature $R$ through 1$\times$1 convolutional layer followed by non-liner ReLU activation, \ieno, $R^+ = ReLU(W^+ R)$, $R^- = ReLU(W^- R)$. \textbf{\emph{SNR${_{g(\cdot)^2}}$}}: we use two unshared channel attention gates $g(\cdot)^+$, $g(\cdot)^-$ to obtain $R^+$ and $R^-$ respectively. \textbf{\emph{SNR-S}}: different from the original SNR design that leverages channel attention to achieve feature separation, here we disentangle the residual feature $R$ using only a spatial attention, and its complementary. \textbf{\emph{SNR-SC}}: we disentangle the residual feature $R$ through the paralleled spatial and channel attention. Table \ref{tab:disentangle} shows the results. We have the following observations: 

\noindent\textbf{1)} Our \textbf{\emph{SNR}} outperforms \textbf{\emph{SNR${_{conv}}$}} by \textbf{1.3\%} on average on PACS, demonstrating the benefit of explicit design of decomposition using attention masks. 

\noindent\textbf{2)} Ours \textbf{\emph{SNR}} outperforms \textbf{\emph{SNR$_{g(\cdot)^2}$}} by \textbf{0.9\%} on average on PACS, demonstrating the benefit of the design that encourages interaction between $R^+$ and $R^-$ where their sum is equal to $R$.

\noindent\textbf{3)} \textbf{\emph{SNR-S}} is inferior to \emph{SNR} that is based on channel attention. Those task-irrelevant style factors (\egno, illumination, contrast, saturation) are in general spatial consistent, which are characterized by the statistics of each channel. IN reduces style discrepancy of input features by performing normalization across spatial dimensions independently for each channel, where the normalization parameters are the same across different spatial positions. Consistent with IN, we disentangle the
features at channel level and add the task-relevant ones back to the normalized features.

\noindent\textbf{4)} \textbf{\emph{SNR-SC}} outperforms \textbf{\emph{SNR}} which uses only channel attention by \textbf{0.3\%} on average on PACS. To be simple and align with our main purpose of distilling the removed task-relevant information, we use only channel attention by default. 
\begin{table}[t]
  \centering
  \caption{Study on the disentanglement designs in SNR}
  \setlength{\tabcolsep}{4mm}{
    \begin{tabular}{c|ccccc}
    \toprule
    \multirow{2}[3]{*}{Method} & \multicolumn{5}{c}{PACS} \\
\cline{2-6}   & \textit{Art}   & \textit{Cat}   & \textit{Pho}   & \textit{Skt}   & \textit{Avg} \\
    \hline
    Baseline (AGG)   & 77.0  & 75.9  & \textbf{96.0} & 69.2  & 79.5 \\
    \hline
    SNR${_{conv}}$ & 77.9  & 76.4  & 95.7  & 71.8  & 80.5 \\
    SNR$_{g(\cdot)^2}$ & 78.7  & 76.9  & 95.2  & 72.8  & 80.9 \\
    SNR   & 80.3  & \textbf{78.2} & 94.5  & 74.1  & 81.8 \\
    {SNR-S}   & 80.1  & 77.9 & 94.0  & 73.6  & 81.4 \\
    SNR-SC & \textbf{80.7} & 77.8  & 94.9  & \textbf{74.8} & \textbf{82.1} \\
    \bottomrule
    \end{tabular}}%
  \label{tab:disentangle}%
\end{table}%

\begin{figure} [t]
  \centerline{\includegraphics[width=0.85\linewidth]{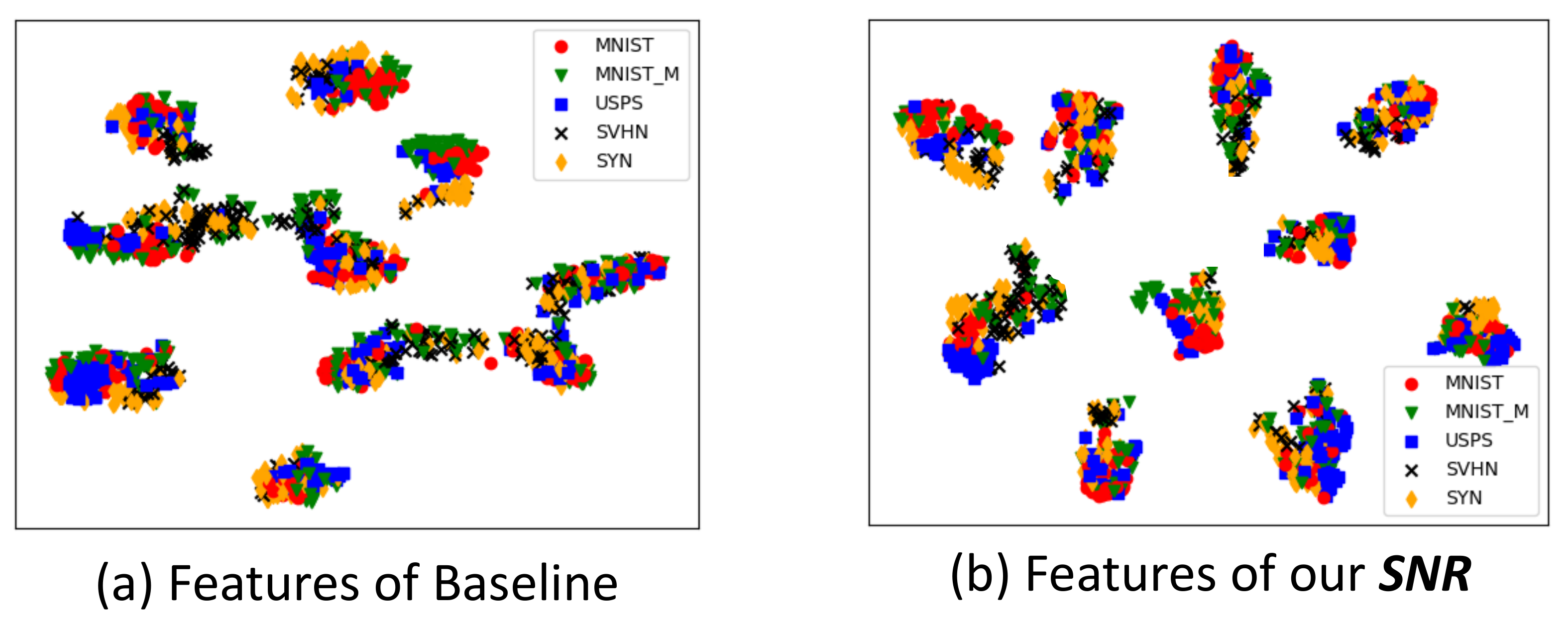}}
    \caption{Visualization of t-SNE distributions on the Digit-Five dataset for UDA classification task. We compare our \emph{SNR-M3SDA} with the baseline scheme \emph{Baseline (M3SDA)}.}
\label{fig:tSNE}
\vspace{-4mm}
\end{figure}
\subsubsection{Visualization} \label{subsec:visualization}

\noindent\textbf{Feature Map Visualization.} To better understand how our SNR works, we visualize the intermediate feature maps of the SNR module that is inserted in the third residual block (\ieno, SNR-3). Following \cite{zheng2011person,zhou2019omni}, we get each activation map by summarizing the feature maps along channels followed by a spatial $\ell_2$ normalization.

Fig.~\ref{fig:vis_ftp}(a) shows the activation maps of normalized feature $\widetilde{F}$, enhanced feature $\widetilde{F}^+=\widetilde{F}+R^+$, and contaminated feature $\widetilde{F}^-=\widetilde{F}+R^-$, respectively. We see that after adding the task-irrelevant feature $R^{-}$, the contaminated feature $\widetilde{F}^-$ has high response mainly on background. In contrast, the enhanced feature $\widetilde{F}^+$ (with the restitution of task-relevant feature $R^{+}$) has high responses on regions of the object (`dog' and `horse'), better capturing discriminative feature regions. 

Moreover, in Fig.~\ref{fig:vis_ftp}(b), we further compare the activation maps $\widetilde{F}^+$ of our scheme and those of the strong baseline scheme \emph{Baseline (AGG)} by varying the styles of input images (\egno, contrast, illumination). We can see that, for the images with different styles, the activation maps of our scheme are more consistent than those of the baseline scheme \emph{Baseline (AGG)}. The activation maps of \emph{Baseline (AGG)} are more disorganized and are easily affected by style variants. These indicate that our scheme is more robust to style variations.


\noindent\textbf{Visualization of Feature Distributions.} 
In Fig.~\ref{fig:tSNE}, we visualize the distribution of the features using t-SNE \cite{maaten2008visualizing} for UDA classification on Digit-Five (on the setting \emph{mm,mt,sv,syn$\rightarrow$up}). We compare the feature distribution of (a) the baseline scheme \emph{Baseline (M3SDA~\cite{peng2019moment})}, and (b) our \emph{SNR}. We observe that the features obtained by our \emph{SNR} are better separated for different classes than the baseline scheme. 

\begin{table*}[t]
  \centering
  \scriptsize
  \caption{Domain generalization performance (\%) for semantic segmentation when we train on GTA5 and test on Cityscapes.}
  \setlength{\tabcolsep}{0.7mm}{
    \begin{tabular}{c|c|c|cccccccccccccccccccc}
    \toprule
    \multicolumn{23}{c}{GTA5$\rightarrow$Cityscape} \\
    \midrule
    Setting & Backbone & Method & \begin{sideways}mIoU\end{sideways} & \begin{sideways}road\end{sideways} & \begin{sideways}sidewalk\end{sideways} & \begin{sideways}building\end{sideways} & \begin{sideways}wall\end{sideways} & \begin{sideways}fence\end{sideways} & \begin{sideways}pole\end{sideways} & \begin{sideways}light\end{sideways} & \begin{sideways}sign\end{sideways} & \begin{sideways}vegetation\end{sideways} & \begin{sideways}terrain\end{sideways} & \begin{sideways}sky\end{sideways} & \begin{sideways}person\end{sideways} & \begin{sideways}rider\end{sideways} & \begin{sideways}car\end{sideways} & \begin{sideways}truck\end{sideways} & \begin{sideways}bus\end{sideways} & \begin{sideways}train\end{sideways} & \begin{sideways}motocycle\end{sideways} & \begin{sideways}bicycle\end{sideways} \\
    \midrule
    \multicolumn{1}{c|}{\multirow{6}[4]{*}{Source\_only}} & \multirow{3}[2]{*}{DRN-D-105} & Baseline & 29.84 & 45.82 & 20.80 & 58.86 & 5.14  & \textbf{16.74} & \textbf{31.74} & \textbf{33.70} & \textbf{19.34} & 83.25 & 15.11 & 66.99 & 52.99 & 9.20  & 53.59 & 12.99 & 14.24 & 3.46  & 17.54 & 5.50 \\
          &       & Baseline-IN & 32.64 & 59.27 & 16.25 & 71.58 & 12.66 & 16.04 & 23.61 & 24.72 & 14.01 & \textbf{84.43} & 31.96 & 62.76 & 52.33 & \textbf{11.34} & 61.00 & \textbf{15.27} & \textbf{21.98} & \textbf{7.43} & 20.48 & 13.07 \\
          &       & \textbf{SNR} & \textbf{36.16} & \textbf{83.34} & \textbf{17.32} & \textbf{78.74} & \textbf{16.85} & 10.71 & 29.17 & 30.46 & 13.76 & 83.42 & \textbf{34.43} & \textbf{73.30} & \textbf{53.95} & 8.95  & \textbf{78.84} & 13.86 & 15.18 & 3.96  & \textbf{21.48} & \textbf{19.39} \\
\cline{2-23}          & \multirow{3}[2]{*}{DeeplabV2} & Baseline & 36.94 & 71.41 & 15.33 & 74.04 & 21.13 & 14.49 & 22.86 & 33.93 & 18.62 & 80.75 & 20.98 & 68.58 & 56.62 & \textbf{27.17} & 67.47 & 32.81 & 5.60  & \textbf{7.74} & 28.43 & 33.82 \\
          &       & Baseline-IN & 39.46 & 73.43 & 22.19 & 78.71 & 24.04 & 15.29 & 27.63 & 29.66 & 19.96 & 80.19 & 27.42 & 70.26 & 56.27 & 15.86 & 72.97 & \textbf{33.66} & 37.79 & 5.63  & 29.20 & 29.59 \\
          &       & \textbf{SNR} & \textbf{42.68} & \textbf{78.95} & \textbf{29.51} & \textbf{79.92} & \textbf{25.01} & \textbf{20.32} & \textbf{28.33} & \textbf{34.83} & \textbf{20.40} & \textbf{82.76} & \textbf{36.13} & \textbf{71.47} & \textbf{59.19} & 21.62 & \textbf{75.84} & 32.78 & \textbf{45.48} & 2.97  & \textbf{30.26} & \textbf{35.13} \\
    \bottomrule
    \end{tabular}}%
  \label{tab:dg_seg}%
\end{table*}%

\begin{table*}[t]
  \centering
  \scriptsize
  \caption{Domain generalization performance (\%) of semantic segmentation when we train on Synthia and test on Cityscapes.}
  \setlength{\tabcolsep}{1mm}{
    \begin{tabular}{c|c|c|ccccccccccccccccc}
    \toprule
    \multicolumn{19}{c}{Synthia$\rightarrow$Cityscape}               &  \\
    \midrule
    Setting & Backbone & Method & \begin{sideways}mIoU\end{sideways} & \begin{sideways}road\end{sideways} & \begin{sideways}sidewalk\end{sideways} & \begin{sideways}building\end{sideways} & \begin{sideways}wall\end{sideways} & \begin{sideways}fence\end{sideways} & \begin{sideways}pole\end{sideways} & \begin{sideways}light\end{sideways} & \begin{sideways}sign\end{sideways} & \begin{sideways}vegetation\end{sideways} & \begin{sideways}sky\end{sideways} & \begin{sideways}person\end{sideways} & \begin{sideways}rider\end{sideways} & \begin{sideways}car\end{sideways} & \begin{sideways}bus\end{sideways} & \begin{sideways}motocycle\end{sideways} & \begin{sideways}bicycle\end{sideways} \\
    \midrule
    \multicolumn{1}{c|}{\multirow{6}[4]{*}{Source\_only}} & \multirow{3}[2]{*}{DRN-D-105} & Baseline & 23.56 & 14.63 & 11.49 & 58.96 & \textbf{3.21} & \textbf{0.10} & 23.80 & 1.32  & 7.20  & 68.49 & 76.12 & \textbf{54.31} & 6.98  & 34.21 & \textbf{15.32} & 0.81  & 0.00 \\
          &       & Baseline-IN & 24.71 & 15.89 & 13.85 & \textbf{63.22} & 2.98  & 0.00  & 26.20 & 2.56  & 8.10  & 70.08 & 77.52 & 53.90 & 7.98  & 35.62 & 15.08 & 2.36  & 0.00 \\
          &       & \textbf{SNR} & \textbf{26.30} & \textbf{19.33} & \textbf{15.21} & 62.54 & 3.07  & 0.00  & \textbf{29.15} & \textbf{6.32} & \textbf{10.20} & \textbf{73.22} & \textbf{79.62} & 53.67 & \textbf{8.92} & \textbf{41.08} & 15.16 & \textbf{3.23} & 0.00 \\
\cline{2-20}          & \multirow{3}[2]{*}{DeeplabV2} & Baseline & 31.12 & 35.79 & 17.12 & 72.29 & 4.51  & 0.15  & 26.52 & 5.76  & 8.23  & 74.94 & 80.71 & \textbf{56.18} & 16.36 & 39.31 & \textbf{21.57} & 10.52 & 27.95 \\
          &       & Baseline-IN & 32.93 & 45.55 & 23.63 & 71.68 & 4.51  & \textbf{0.42} & 29.36 & \textbf{12.52} & \textbf{14.34} & 74.94 & 80.96 & 50.53 & \textbf{20.15} & 42.41 & 11.20 & 10.30 & \textbf{34.45} \\
          &       & \textbf{SNR} & \textbf{34.36} & \textbf{50.43} & \textbf{23.64} & \textbf{74.41} & \textbf{5.82} & 0.37  & \textbf{30.37} & 12.24 & 13.52 & \textbf{78.35} & \textbf{83.05} & 55.29 & 18.13 & \textbf{47.10} & 13.73 & \textbf{12.64} & 30.70 \\
    \bottomrule
    \end{tabular}}%
  \label{tab:dg_seg_2}%
  \vspace{-3mm}
\end{table*}%

\subsection{Semantic Segmentation}\label{subsec:seg}

\subsubsection{Datasets and Implementation Details}

For the semantic segmentation task, we used three representative semantic segmentation datasets: Cityscapes \cite{Cordts2016Cityscapes}, Synthia \cite{ros2016synthia}, and GTA5 \cite{Richter_2016_ECCV}. The detailed datasets introduction and experimental implementation details can be found in \textbf{Supplementary}.

\subsubsection{Results on Domain Generalization}

Here, we evaluate the effectiveness of SNR under DG setting (only training on the source datasets, and directly testing on the target test set). Since very few previous works investigate on this task, here we define the comparison/validation settings. We compare the proposed scheme \emph{SNR} with 1) the \emph{Baseline} (only use source dataset for training) and 2) the baseline when adding IN after each convolutional block \emph{Baseline-IN}.




Table \ref{tab:dg_seg} and Table \ref{tab:dg_seg_2} show that for DRN-D-105, our scheme \emph{SNR} outperforms \emph{Baseline} by 6.32\% and 2.74\% in mIoU accuracy for GTA5-to-Cityscapes and Synthia-to-Cityscapes, respectively. For the stronger backbone network DeeplabV2, our scheme \emph{SNR}  outperforms \emph{Baseline} by 5.74\% and 3.24\% in mIoU for GTA5-to-Cityscapes and Synthia-to-Cityscapes, respectively. When compared with the scheme \emph{Baseline-IN}, our \emph{SNR} also consistently outperforms it on two backbones for both settings.


\begin{table*}[htbp]
  \centering
  \caption{Performance (\%) comparisons with the state-of-the-art semantic segmentation approaches for unsupervised domain adaptation for GTA5-to-Cityscape.}
  \setlength{\tabcolsep}{0.8mm}{
    \begin{tabular}{c|p{24.78em}|c|ccccccccccccccccccc}
    \toprule
    \multicolumn{22}{c}{GTA5$\rightarrow$Cityscape} \\
    \midrule
    Network & \multicolumn{1}{c|}{method} & \begin{sideways}mIoU\end{sideways} & \begin{sideways}road\end{sideways} & \begin{sideways}sdwk\end{sideways} & \begin{sideways}bldng\end{sideways} & \begin{sideways}wall\end{sideways} & \begin{sideways}fence\end{sideways} & \begin{sideways}pole\end{sideways} & \begin{sideways}light\end{sideways} & \begin{sideways}sign\end{sideways} & \begin{sideways}vgttn\end{sideways} & \begin{sideways}trrn\end{sideways} & \begin{sideways}sky\end{sideways} & \begin{sideways}person\end{sideways} & \begin{sideways}rider\end{sideways} & \begin{sideways}car\end{sideways} & \begin{sideways}truck\end{sideways} & \begin{sideways}bus\end{sideways} & \begin{sideways}train\end{sideways} & \begin{sideways}mcycl\end{sideways} & \begin{sideways}bcycl\end{sideways} \\
    \hline
    \multirow{3}[3]{*}{DRN-105} 
          & \multicolumn{1}{c|}{DANN  ~\cite{ganin2016domain}} & 32.8  & 64.3  & 23.2  & 73.4  & 11.3  & 18.6  & 29.0  & 31.8  & 14.9  & 82.0  & 16.8  & 73.2  & 53.9  & 12.4  & 53.3  & 20.4  & 11.0  & 5.0   & 18.7  & 9.8 \\
          & \multicolumn{1}{c|}{MCD ~\cite{saito2018maximum}} & 35.0 & 87.5  & 17.6  & 79.7  & 22.0  & 10.5  & 27.5  & 21.9  & 10.6  & 82.7  & 30.3  & 78.2  & 41.1  & 9.7   & 80.4  & 19.3  & 23.1  & 11.7  & 9.3   & 1.1 \\
      & \multicolumn{1}{c|}{\textbf{SNR-MCD (ours)}} & {\textbf{40.3}} & 87.7  & 36.0  & 80.0  & 19.7  & 19.1  & 30.9  & 32.4  & 13.0  & 82.8  & 34.9  & 79.1  & 50.3  & 11.0  & 84.3  & 23.0  & 28.6  & 16.8  & 18.5  & 17.9 \\
    \hline
    \multirow{4}[3]{*}{DeeplabV2} 
          & \multicolumn{1}{c|}{AdaptSegNet ~\cite{tsai2018learning}} & 42.4  & 86.5  & 36.0  & 79.9  & 23.4  & 23.3  & 23.9  & 35.2  & 14.8  & 83.4  & 33.3  & 75.6  & 58.5  & 27.6  & 73.7  & 32.5  & 35.4  & 3.9   & 30.1  & 28.1 \\
          & \multicolumn{1}{c|}{MinEnt ~\cite{vu2019advent}} & 42.3  & 86.2  & 18.6  & 80.3  & 27.2  & 24.0  & 23.4  & 33.5  & 24.7  & 83.3  & 31.0  & 75.6  & 54.6  & 25.6  & 85.2  & 30.0  & 10.9  & 0.1   & 21.9  & 37.1 \\
          & \multicolumn{1}{c|}{AdvEnt+MinEnt ~\cite{vu2019advent}} & 44.8  & 87.6  & 21.4  & 82.0  & 34.8  & 26.2  & 28.5  & 35.6  & 23.0  & 84.5  & 35.1  & 76.2  & 58.6  & 30.7  & 84.8  & 34.2  & 43.4  & 0.4   & 28.4  & 35.3 \\
          & \multicolumn{1}{c|}{MaxSquare (MS) ~\cite{chen2019domain}} & 44.3  & 88.1  & 27.7  & 80.8  & 28.7  & 19.8  & 24.9  & 34.0  & 17.8  & 83.6  & 34.7  & 76.0  & 58.6  & 28.6  & 84.1  & 37.8  & 43.1  & 7.2   & 32.2  & 34.2 \\
         & \multicolumn{1}{c|}{\textbf{SNR-MS (ours)}} & {\textbf{46.5}} & 90.8  & 40.9  & 81.6  & 29.8  & 23.5  & 24.4  & 34.1  & 21.6  & 84.0  & 39.6  & 77.0  & 59.3  & 30.9  & 84.4  & 37.8  & 44.6  & 8.5   & 33.2  & 37.9 \\
    \bottomrule
    \end{tabular}}%
  \label{tab:uda_seg_1}%
\end{table*}%

\begin{table*}[htbp]
  \centering
  \caption{Performance (\%) comparisons with the state-of-the-art semantic segmentation approaches for unsupervised domain adaptation for Synthia-to-Cityscape.}
  \setlength{\tabcolsep}{1.0mm}{
    \begin{tabular}{c|p{24.78em}|c|cccccccccccccccc}
    \toprule
    \multicolumn{19}{c}{Synthia$\rightarrow$Cityscape} \\
    \hline
    Network & \multicolumn{1}{c|}{method} & \begin{sideways}mIoU\end{sideways} & \begin{sideways}road\end{sideways} & \begin{sideways}sdwk\end{sideways} & \begin{sideways}bldng\end{sideways} & \begin{sideways}wall\end{sideways} & \begin{sideways}fence\end{sideways} & \begin{sideways}pole\end{sideways} & \begin{sideways}light\end{sideways} & \begin{sideways}sign\end{sideways} & \begin{sideways}vgttn\end{sideways} & \begin{sideways}trrn\end{sideways} & \begin{sideways}sky\end{sideways} & \begin{sideways}person\end{sideways} & \begin{sideways}car\end{sideways} & \begin{sideways}bus\end{sideways} & \begin{sideways}mcycl\end{sideways} & \begin{sideways}bcycl\end{sideways} \\
    \midrule
    \multirow{3}[3]{*}{DRN-105} 
          & \multicolumn{1}{c|}{DANN  ~\cite{ganin2016domain}} & 32.5  & 67.0  & 29.1  & 71.5  & 14.3  & 0.1   & 28.1  & 12.6  & 10.3  & 72.7  & 76.7  & 48.3  & 12.7  & 62.5  & 11.3  & 2.7   & 0.0 \\
          & \multicolumn{1}{c|}{MCD ~\cite{saito2018maximum}} & 36.6  & 84.5  & 43.2  & 77.6  & 6.0   & 0.1   & 29.1  & 7.2   & 5.6   & 83.8  & 83.5  & 51.5  & 11.8  & 76.5  & 19.9  & 4.7   & 0.0 \\
            & \multicolumn{1}{c|}{\textbf{SNR-MCD (ours)}} & {\textbf{39.6}} & 88.1  & 55.4  & 71.7  & 16.3  & 0.2   & 27.6  & 13.0  & 11.3  & 82.4  & 82.0  & 55.0  & 13.7  & 83.3  & 27.8  & 6.7   & 0.0 \\
    \hline
    \multirow{4}[3]{*}{DeeplabV2} 
          & \multicolumn{1}{c|}{AdaptSegNet ~\cite{tsai2018learning}} & -     & 84.3  & 42.7  & 77.5  & -     & -     & -     & 4.7   & 7.0   & 77.9  & 82.5  & 54.3  & 21.0  & 72.3  & 32.2  & 18.9  & 32.3 \\
          & \multicolumn{1}{c|}{MinEnt ~\cite{vu2019advent}} & 38.1  & 73.5  & 29.2  & 77.1  & 7.7   & 0.2   & 27.0  & 7.1   & 11.4  & 76.7  & 82.1  & 57.2  & 21.3  & 69.4  & 29.2  & 12.9  & 27.9 \\
          & \multicolumn{1}{c|}{AdvEnt+MinEnt ~\cite{vu2019advent}} & 41.2  & 85.6  & 42.2  & 79.7  & 8.7   & 0.4   & 25.9  & 5.4   & 8.1   & 80.4  & 84.1  & 57.9  & 23.8  & 73.3  & 36.4  & 14.2  & 33.0 \\
          & \multicolumn{1}{c|}{MaxSquare (MS) ~\cite{chen2019domain}} & 39.3  & 77.4  & 34.0  & 78.7  & 5.6   & 0.2   & 27.7  & 5.8   & 9.8   & 80.7  & 83.2  & 58.5  & 20.5  & 74.1  & 32.1  & 11.0  & 29.9 \\
          & \multicolumn{1}{c|}{\textbf{SNR-MS (ours)}} & {\textbf{45.1}} & 90.0  & 37.1  & 82.0  & 10.3  & 0.9   & 27.4  & 15.1  & 26.3  & 82.9  & 76.6  & 60.5  & 26.6  & 86.0  & 41.3  & 31.6  & 27.6 \\
    \bottomrule
    \end{tabular}}%
  \label{tab:uda_seg_2}%
\end{table*}%

\subsubsection{Results on Unsupervised Domain Adaptation}

Unsupervised domain adaptive semantic segmentation has been extensively studied \cite{saito2018maximum,chen2019domain}, where the unlabeled target domain data is also used for training. We validate the effectiveness of  UDA by adding the SNR modules into two popular UDA approaches: MCD~\cite{saito2018maximum} and MaxSqure(MS)~\cite{chen2019domain}, respectively. MCD \cite{saito2018maximum} maximizes the discrepancy between two task-classifiers while minimizing it with respect to the feature extractor of domain adaptation. MS ~\cite{chen2019domain} extends the entropy
minimization idea to UDA for semantic segmentation by using a proposed maximum squares loss. We refer to the two schemes powered by our SNR modules as \emph{SNR-MCD} and \emph{SNR-MS}. Table \ref{tab:uda_seg_1} and Table \ref{tab:uda_seg_2} show that based on the same DRN-105 backbone, \emph{SNR-MCD} significantly outperforms the second-best method MCD \cite{saito2018maximum} by \textbf{5.3\%}, and \textbf{3.0\%} in mIoU for GTA5$\rightarrow$Cityscape and Synthia$\rightarrow$Cityscape, respectively. In addition, based on the DeeplabV2 backbone, \emph{SNR-MS} consistently outperforms \emph{MaxSqure~(MS)} \cite{chen2019domain} by \textbf{2.2\%}, and \textbf{5.8\%} in mIoU for GTA5$\rightarrow$Cityscape and Synthia$\rightarrow$Cityscape, respectively.



\begin{figure}[t]
  \centerline{\includegraphics[width=1.0\linewidth]{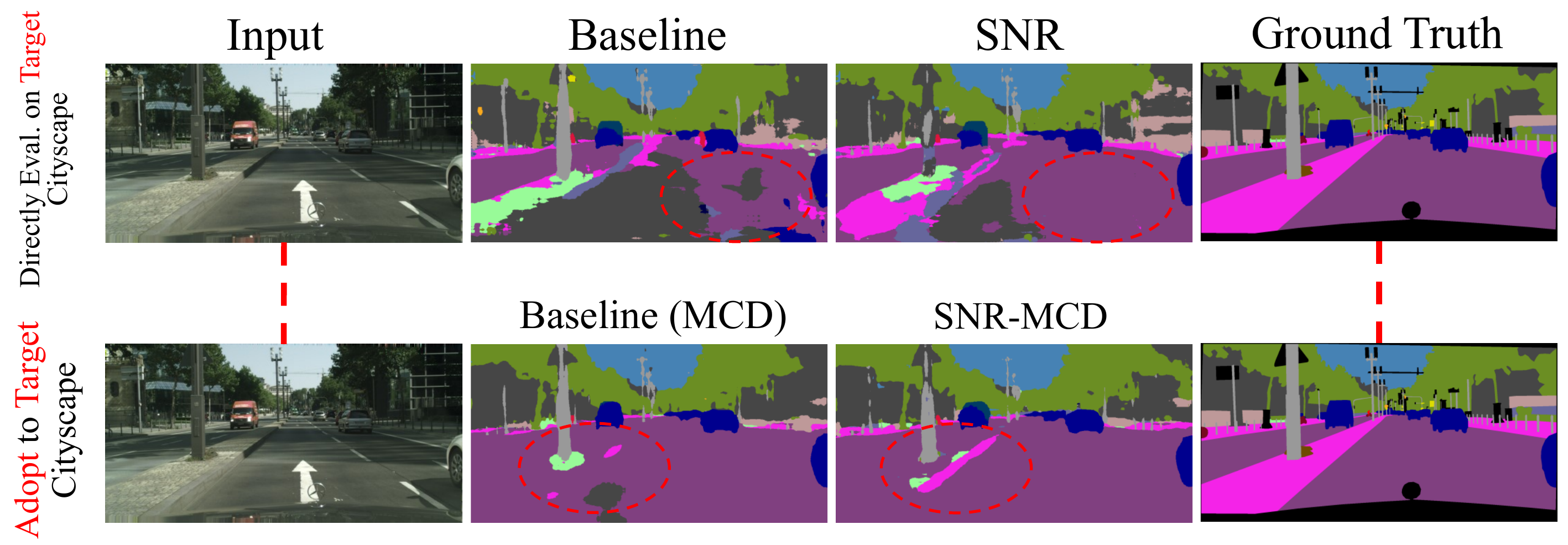}}
    \caption{Qualitative results on domain generable segmentation (first row) and domain adaptive segmentation (second row) from GTA5 to Cityscapes. For DG (first row), \emph{Baseline} denotes the baseline scheme trained with source domain dataset while testing on the target domain directly. \emph{SNR} denotes our scheme which adds SNR modules to \emph{Baseline}. For UDA (second row), we compare the baseline scheme \emph{Baseline (MCD)}~\cite{saito2018maximum} to the scheme \emph{SNR+MCD} which is powered by our SNR.}
\label{fig:seg}
\vspace{-5mm}
\end{figure}

\subsubsection{Visualization of DG and UDA Results}
We visualize the qualitative results in Fig.~\ref{fig:seg} by comparing the baseline schemes and the schemes powered by our SNR. For DG in the first row, we can see that the introduction of SNR to \emph{Baseline} brings obvious improvement on the segmentation results. For UDA in the second row, 1) the introduction of SNR to \emph{Baseline (MCD)} brings clear improvement on the segmentation results; 2) the segmentation results with adapation (UDA) to the target domain data is much better than that obtained from domain generalization model, indicating the exploration of target domain data is helpful to have good performance. 

\subsection{Object Detection}
\label{subsec:det}

\subsubsection{Datasets and Implementation Details}

Following \cite{chen2018domain,khodabandeh2019robust}, we evaluate performance on multi- and single-label object detection tasks using three different datasets: Cityscapes~\cite{Cordts2016Cityscapes}, Foggy Cityscapes~\cite{sakaridis2018semantic}, and KITTI~\cite{Geiger2013IJRR}. The detailed datasets introduction can be found in \textbf{Supplementary}.

\begin{table*}[htbp]
  \centering
  \caption{Performance (in mAP accuracy \%) of object detection on the Foggy Cityscapes validation set, models are trained on the Cityscapes training set.}
    \begin{tabular}{c|c|cccccccc|c}
    \toprule
    \multirow{2}[4]{*}{Setting} & \multirow{2}[4]{*}{Method} & \multicolumn{9}{c}{Cityscapes$\rightarrow$Foggy Cityscapes} \\
\cline{3-11}          &       & person & rider & car   & truck & bus   & train & mcycle & bicycle & mAP \\
    \hline
    \multirow{2}[1]{*}{DG} & Faster R-CNN~\cite{ren2015faster} & 17.8  & 23.6  & 27.1  & 11.9  & 23.8  & 9.1   & 14.4  & 22.8  & 18.8 \\
          & SNR-Faster R-CNN   & \textbf{20.3} & \textbf{24.6} & \textbf{33.6} & \textbf{15.9} & \textbf{26.3} & \textbf{14.4} & \textbf{16.8} & \textbf{26.8} & \textbf{22.3} \\
    \hline
    \multirow{2}[1]{*}{UDA} & DA Faster R-CNN~\cite{chen2018domain} & 25.0  & 31.0  & 40.5  & 22.1  & 35.3  & 20.2  & 20.0  & 27.1  & 27.6 \\
          & SNR-DA Faster R-CNN   & \textbf{27.3} & \textbf{34.6} & \textbf{44.6} & \textbf{23.9} & \textbf{38.1} & \textbf{25.4} & \textbf{21.3} & \textbf{29.7} & \textbf{30.6} \\
    \bottomrule
    \end{tabular}%
  \label{tab:detec_1}%
  \vspace{-2mm}
\end{table*}%

\begin{table}[htbp]
  \centering
  \caption{Performance (in AP accuracy \%) for the class of Car for object detection on KITTI (K) and Cityscapes (C).}
  \setlength{\tabcolsep}{5.0mm}{
    \begin{tabular}{c|c|cc}
    \toprule
    \multirow{2}[2]{*}{Setting} & \multirow{2}[2]{*}{Method} & \multirow{2}[2]{*}{K$\rightarrow$C} & \multirow{2}[2]{*}{C$\rightarrow$K} \\
          &       &       &  \\
    \hline
    \multirow{2}[1]{*}{DG} & Faster R-CNN~\cite{ren2015faster} & 30.24 & 53.52 \\
          & SNR   & \textbf{35.92} & \textbf{57.94} \\
    \hline
    \multirow{2}[1]{*}{UDA} & DA Faster R-CNN~\cite{chen2018domain} & 38.52 & 64.15 \\
          & SNR   & \textbf{43.51} & \textbf{69.17} \\
    \bottomrule
    \end{tabular}}%
  \label{tab:detec_2}%
  \vspace{-2mm}
\end{table}%

\begin{figure}[htbp]
\centerline{\includegraphics[width=0.9\linewidth]{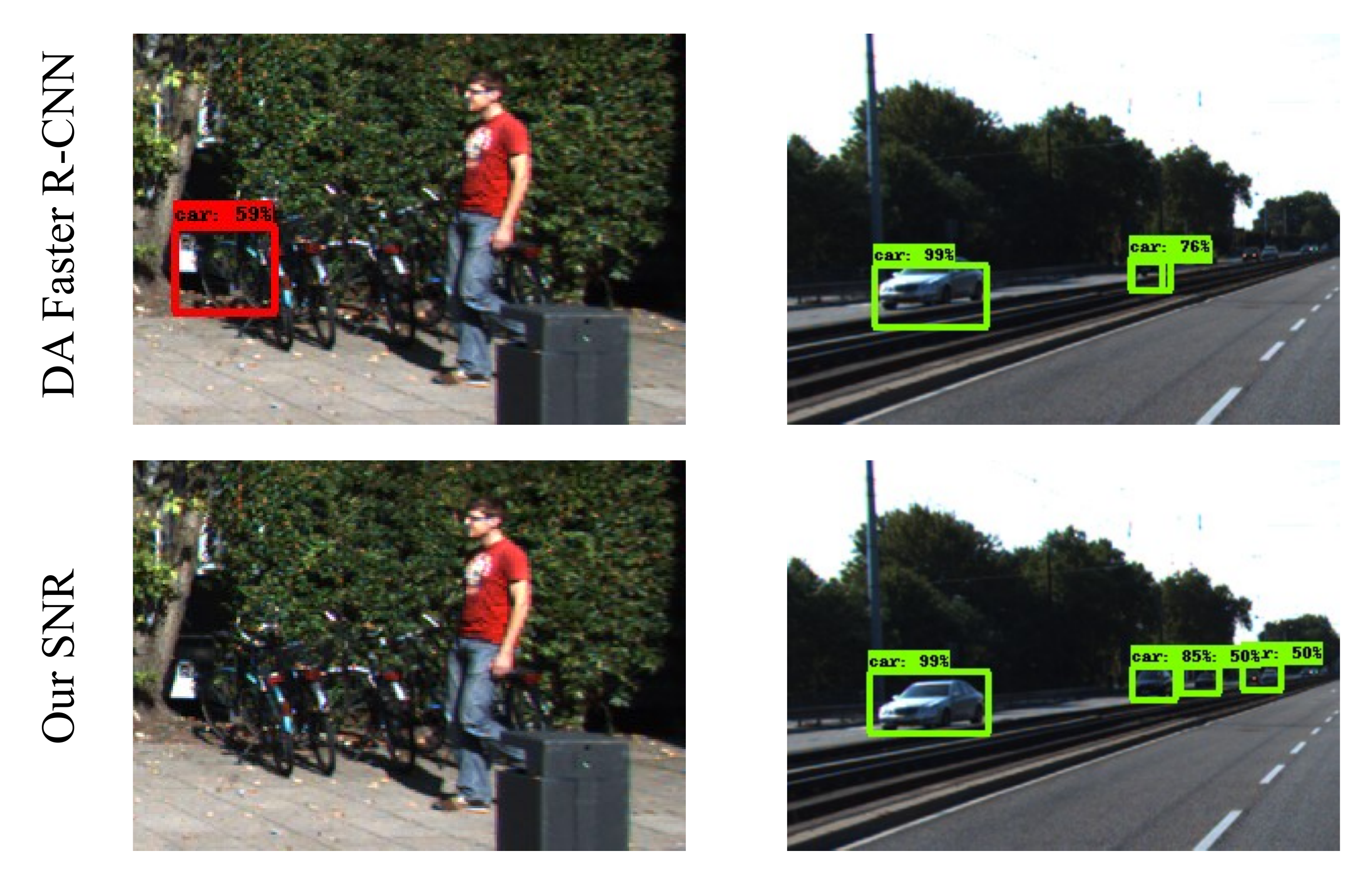}}
\caption{Qualitative comparisons of the baseline approach DA Faster R-CNN ~\cite{chen2018domain} and the baseline powered by our SNR on  ``\city{} $\rightarrow$ \kit{}''. Top and bottom rows denote the detected cars by the baseline scheme \emph{DA Faster R-CNN} and our scheme \emph{SNR-DA Faster R-CNN} respectively. }
\label{fig:det}
\end{figure}

\begin{table}[htbp]
  \centering
  \caption{Comparisons of complexity and model sizes. FLOPs: the number of FLoating-point OPerations; Params: the number of parameter.}
  \setlength{\tabcolsep}{7mm}{
    \begin{tabular}{l|cc}
    \toprule
          & FLOPs & Params \\
    \midrule
    ResNet-18 & 1.83G & 11.74M \\
    ResNet-18-SNR & 2.03G & 12.30M \\
    $\Delta$  & +9.80\% & +4.50\% \\
    \midrule
    \midrule
    ResNet-50 & 3.87G & 24.56M \\
    ResNet-50-SNR & 4.08G & 25.12M \\
    $\Delta$  & +5.10\% & +2.20\% \\
    \bottomrule
    \end{tabular}}%
  \label{tab:complexity}%
  \vspace{-4mm}
\end{table}%

For the domain generalization (DG) experiments, we employ the original Faster RCNN \cite{ren2015faster} as our baseline, which is trained using the source domain training data. We follow \cite{ren2015faster} to set the hyper-parameters. For our scheme \emph{SNR}, we add SNR modules into the backbone (by adding a SNR module after each convolutional block for the first four blocks of ResNet-50) of the Faster RCNN, which are initialized using weights pre-trained on ImageNet. We train the network with a learning rate of $0.001$ for $50$k iterations and then reduce the learning rate to $0.0001$ for another $20$k iterations.

For the unsupervised domain adaptation (UDA) experiments, we use the Domain Adaptive Faster R-CNN (Da Faster R-CNN)~\cite{chen2018domain} model as our baseline, which tackles the domain shift on two levels, the image level and the instance level. A domain classifier is added on each level, trained in an adversarial training manner. A consistency regularizer is incorporated
within these two classifiers to learn a domain-invariant RPN for the Faster R-CNN model. Each batch is composed
of two images, one from the source domain and the other from the
target domain. A momentum of 0.9 and a weight decay of
0.0005 is used in our experiments.

For all experiments \footnote{We use the repository https://github.com/yuhuayc/da-faster-rcnn-domain-adaptive-faster-r-cnn-for-object-detection-in-the-wild as our code base.}, we report mean average precisions (mAP) with a threshold of 0.5 for evaluation.


\subsubsection{Results on DG and UDA}

\textbf{Results for Normal to Foggy Weather.} Differences in weather conditions can significantly affect visual data. In many applications (\ieno, autonomous driving), the object detector needs to perform well in all conditions~\cite{sakaridis2018semantic}. Here we evaluate the effectiveness of our SNR and demonstrate its generalization superiority over the current state-of-the-art for this task. We use \city{} dataset as the source domain and \cityfog{} as the target domain (denoted by ``\city{} $\rightarrow$ \cityfog{}''). 

Table~\ref{tab:detec_1} compares our schemes using SNR to two baselines (Faster R-CNN~\cite{ren2015faster}, and Domain Adaptive (DA) Faster R-CNN~\cite{chen2018domain}) on domain generalization, and domain adaptation settings. We report the average precision for each category, and the mean average precision (mAP) of all the objects. We can see that our SNR improves Faster R-CNN by $3.5\%$ in mAP for domain generalization, and improves DA Faster R-CNN by $3.0\%$ in mAP for unsupervised domain adaptation.



\noindent\textbf{Results for Cross-Dataset DG and UDA.} Many factors could result in domain gaps. There is usually some data bias when collecting the {datasets~\cite{torralba2011unbiased}}. For example, different datasets are usually captured by different cameras or collected by different organizations with different preference, with different image quality/resolution/characteristics. 
In this subsection, we conduct experiments on two datasets: \city{} and \kit{}. We only train the detector on annotated \textit{cars} because \textit{cars} is the only object common to both \city{} and \kit{}. 


Table~\ref{tab:detec_2} compares our methods to two baselines: Faster R-CNN~\cite{ren2015faster}, and Domain Adaptive (DA) Faster R-CNN~\cite{chen2018domain} for domain generalization and domain adaptation setting, respectively. We denote \kit{} (source dataset) to \city{} (target dataset) as $K \rightarrow C$ and vice versa. We can see that the introduction of SNR brings significant performance improvement for both DG and UDA settings. 



\subsubsection{Qualitative Results}

For UDA, we visualize the qualitative detection results in Fig. \ref{fig:det}. We see that our SNR corrects several false positives in the first column, and has detected cars that DA Faster R-CNN missed in the second column.

\subsection{Complexity Analysis}

In Table~\ref{tab:complexity}, we analyze the increase of complexity of our SNR modules in terms of FLOPs and model size with respect to different backbone networks. Here, we use our default setting where we insert a SNR module after each convolutional block (for the first four blocks) for the backbone networks of ResNet-18, ResNet-50. We observe that our SNR modules bring a small increase in complexity. For ResNet-50~\cite{he2016deep} backbone, our SNR only brings an increase of 
2.2\% in model size (24.56M vs. 25.12M) and an increase of 5.1\% in computational complexity (3.87G vs. 4.08G FLOPs).


\section{Conclusion}
In this paper, we present a Style Normalization and Restitution (SNR) module, which aims to learn generalizable and discriminative feature
representations for effective domain generalization and adaptation.  SNR is generic. As a plug-and-play module, it can be inserted into existing backbone networks for many computer vision tasks. SNR reduces the style variations by using Instance Normalization (IN). To prevent the loss of task-relevant discriminative information cased by IN, we propose to distill task-relevant discriminative features from the discarded residual features and add them back to the network, through a well-designed restitution step. Moreover, to promote a better feature disentanglement of task-relevant and task-irrelevant information, we introduce a dual \ourloss constraint. Extensive experimental results demonstrate the effectiveness of our SNR module for both domain generalization and domain adaptation. The schemes powered by SNR achieves the state-of-the-art performance on various tasks, including classification, semantic segmentation, and object detection.


\clearpage


\noindent{\LARGE \textbf{Appendix}}
\vspace{2mm}

\section{Datasets and Implementation Details}

\subsection{Object Classification}

\begin{figure*}[th]
  \centerline{\includegraphics[width=1.0\linewidth]{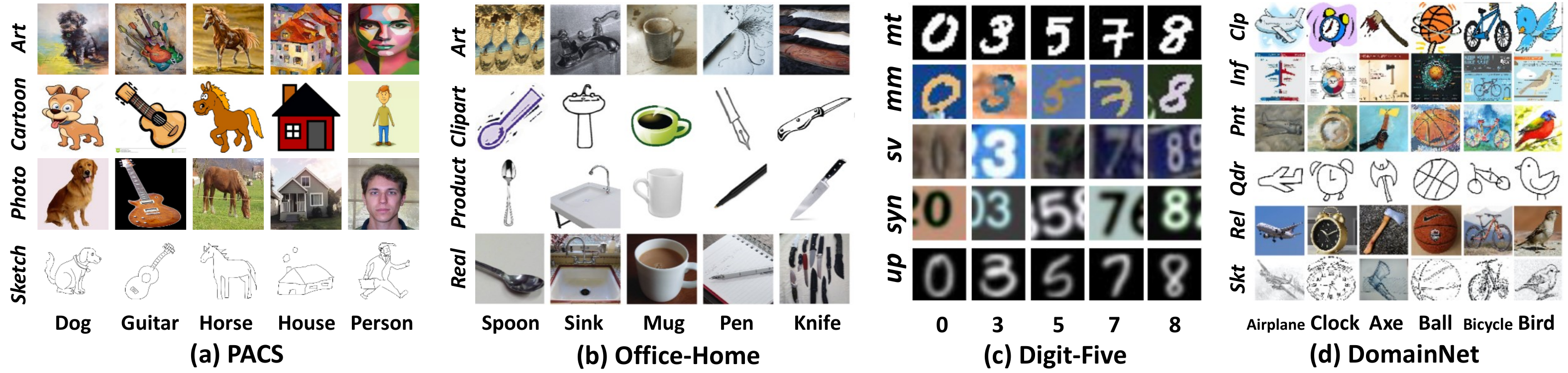}}
  \vspace{-1mm}
    \caption{Four classification datasets (first two for DG and last two for UDA). (a) PACS, which includes \textit{Sketch}, \textit{Photo}, \textit{Cartoon}, and \textit{Art}. (b) Office-Home, which includes {Real-world} (\textit{Real}), \textit{Product}, \textit{Clipart}, and \textit{Art}. (c) Digit-Five, which includes MNIST \cite{lecun1998mnist} (\textit{\textbf{mt}}), MNIST-M \cite{ganin2015unsupervised} (\textit{\textbf{mm}}), USPS \cite{hull1994database} (\textit{\textbf{up}}), SVHN \cite{netzer2011svhn} (\textit{\textbf{sv}}), and Synthetic \cite{ganin2015unsupervised} (\textit{\textbf{syn}}). (d) DomainNet, which includes {Clipart} (\textit{\textbf{clp}}), {Infograph} (\textit{\textbf{inf}}), {Painting} (\textit{\textbf{pnt}}), {Quickdraw} (\textit{\textbf{qdr}}), {Real} (\textit{\textbf{rel}}), and {Sktech} (\textit{\textbf{skt}}). Considering the required huge computation resources, we use a subset of DomainNet (\ieno, mini-DomainNet) following \cite{zhou2020domain} for ablation experiments. The full DomainNet dataset is also used for performance comparison with the state-of-the-art methods.}
\label{fig:datasets}
\vspace{-3mm}
\end{figure*}

Fig.~\ref{fig:datasets} shows some samples of these datasets. PACS~\cite{li2017deeper} and Office-Home~\cite{office_home} are two widely used DG datasets where each dataset includes four domains. PACS has seven object categories and office-Home has 65 categories. Digit-Five consists of five different digit recognition datasets: MNIST \cite{lecun1998mnist}, MNIST-M \cite{ganin2015unsupervised}, USPS \cite{hull1994database}, SVHN \cite{netzer2011svhn} and SYN \cite{ganin2015unsupervised}. We follow the same split setting as \cite{peng2019moment} to use the dataset. DomainNet is a recently introduced benchmark for large-scale multi-source domain adaptation \cite{peng2019moment}, which includes six domains (\ieno, Clipart, Infograph, Painting, Quickdraw, Real, and Sketch) of 600k images (345 classes). Considering the high demand on computational resources, following \cite{zhou2020domain}, we use a subset of DomainNet, \ieno, mini-DomainNet, for ablation experiments. The full DomainNet dataset is also used for performance comparison with the state-of-the-art methods.

PACS and Office-Home are usually used for DG. We validate the effectiveness of  DG on PACS and Office-Home. Following \cite{peng2019moment}, we use the leave-one-domain-out protocol. For PACS and Office-Home\footnote{We use the baseline code from Epi-FCR \cite{li2019episodic} https://github.com/HAHA-DL/Episodic-DG as our code framework to validate the effectiveness of our  PACS and Office-Home.}, similar to \cite{cvpr19JiGen,li2019episodic}, we use ResNet18 as the backbone to build our baseline network. We train the model for 40 epochs with an initial learning rate of 0.002. Each mini-batch contains 30 images (10 per source domain). We insert a SNR module after each convolutional block of the ResNet18 baseline as our \emph{SNR} scheme.

Digit-5 and DomainNet are usually used for DA. We validate the effectiveness of our  DA on them. We follow prior works \cite{li2017deeper,cvpr19JiGen,li2019episodic} to use the leave-one-domain-out protocol. For Digit-5, following~\cite{peng2019moment}, we build the backbone with three convolution layers and two fully connected layers\footnote{We use the baseline code from DEAL \cite{zhou2020domain} https://github.com/KaiyangZhou/Dassl.pytorch as our code framework to validate the effectiveness of  Digit-5 and mini-DomainNet datasets.}. We insert a SNR module after each convolutional layer of the baseline as our \emph{SNR} scheme. For each mini-batch, we sample 64 images from each domain. The model is trained with an initial learning rate of 0.05 for 30 epochs. For mini-DomainNet, we use ResNet18~\cite{he2016deep} as the backbone. For full DomainNet, we use ResNet152~\cite{he2016deep} as the backbone. We insert a SNR module after each convolutional block of the ResNet baseline as our \emph{SNR} scheme. We sample 32 images from each domain to form a mini-batch (of size 32$\times$4=128) and train the model for 60 epochs with an initial learning rate of 0.005.  In all experiments, SGD with momentum is used as the optimizer and a cosine annealing rule~\cite{cosineLR} is adopted for learning rate decay.

\subsection{Semantic Segmentation}

Cityscapes contains 5,000 annotated images with 2048$\times$1024 resolution captured from real urban street scenes. GTA5 contains 24,966 annotated images with 1914$\times$1052 resolution obtained from the GTA5 game. For SYNTHIA, we use the subset SYNTHIA-RAND-CITYSCAPES which consists of 9,400 synthetic images of resolution 1280$\times$760.

Following the prior works \cite{hoffman2016fcns,zhang2017curriculum,saito2018maximum}, we use the labeled training set of GTA5 or SYNTHIA
as the source domain and the Cityscapes validation set as our test set. We adopt the Intersection-over-Union
(IoU) of each class and the mean-Intersection-over-Union
(mIoU) as evaluation metrics. We consider the IoU and mIoU of all the 19 classes in the \emph{GTA5-to-Cityscapes} case. Since SYNTHIA has only 16 shared classes with Cityscapes, we consider the IoU and mIoU of the 16 classes in the \emph{SYNTHIA-to-Cityscapes} setting.

As discussed in \cite{tsai2018learning,chen2019domain}, it is also important to adopt a stronger baseline model to understand the effect of different generalization/adaption approaches and to enhance the performance for the practical applications. Therefore, similar to \cite{saito2018maximum}, in all experiments, we employ two kinds of backbones for evaluation.

\noindent1) We use DRN-D-105 \cite{yu2017dilated,saito2018maximum} as our baseline network and apply our SNR to the network. For DRN-D-105, we follow the implementation of MCD\footnote{https://github.com/mil-tokyo/MCD\_DA/tree/master/segmentation}. Similiar to ResNet~\cite{he2016deep}, DRN still uses the block-based architecture. We insert our SNR module after each convolutional block of DRN-D-105. We use momentum SGD to optimize our models. We set the momentum rate to 0.9 and the learning rate to 10e-3 in all experiments. The image size is resized to 1024$\times$512. Here, we report the output results obtained after 50,000 iterations.

\noindent2) We also use Deeplabv2~\cite{deeplabv2} with ResNet-101~\cite{ResNet} backbone that is pre-trained on ImageNet~\cite{ImageNet} as our baseline network, which is the same as other works~\cite{AdaptSegNet, ADVENT}. We insert our SNR module after each convolutional block of ResNet-101. Following the implementation of MSL~\cite{chen2019domain}\footnote{https://github.com/ZJULearning/MaxSquareLoss}, we train the model with SGD optimizer with the learning rate $2.5 \times 10^{-4}$, momentum $0.9$, and weight decay $5 \times10^{-4}$. We schedule the learning rate using ``poly" policy: the learning rate is multiplied by $(1-\frac{iter}{max\_iter})^{0.9}$~\cite{deeplabv2}. Similar to \cite{zhao2017pyramid}, we employ the random flipping and Gaussian blur for data augmentation.

\subsection{Object Detection}

Cityscapes~\cite{Cordts2016Cityscapes} is a dataset\footnote{This dataset is usually used for semantic segmentation as we described before.} of real urban scenes containing $3,475$ images captured by a dash-cam. $2,975$ images are used for training and the remaining $500$ for validation (such split information is different from the above-mentioned statistics for the semantic segmentation). Following ~\cite{chen2018domain}, we report results on the validation set because we do not have annotations of the test set. There are $8$ different object categories in this dataset including \textit{person, rider, car, truck, bus, train, motorcycle and bicycle}. 
    
Foggy Cityscapes~\cite{sakaridis2018semantic} is the foggy version of Cityscapes. The depth maps provided in Cityscapes are used to simulate three intensity levels of fog in \cite{sakaridis2018semantic}. In our experiments we used the fog level with highest intensity (least visibility) to imitate large domain gap. The same dataset split as used for Cityscapes is used for Foggy Cityscapes. 
    
KITTI~\cite{Geiger2013IJRR} is another real-world dataset consisting of $7,481$ images of real-world traffic situations, including freeways, urban and rural areas. Following~\cite{chen2018domain}, we use the entire dataset for training, when it is used as source. We use the entire dataset for testing when it is used as target test set for DG.


%





\ifCLASSOPTIONcaptionsoff
  \newpage
\fi




\bibliographystyle{IEEEtran}
\bibliography{IEEEabrv}
\end{document}